\documentclass[preprint]{neus2025}
\usepackage[T1]{fontenc}
\author[de las Heras Molins, Yalcinkaya, Peters, Fridovich-Keil, \& Bakirtzis]{
\Name{Pau {de las Heras Molins}}\Email{delasherasmolins@telecom-paris.fr}\\\addr LTCI, Télécom Paris, Institut Polytechnique de Paris \AND
\Name{Beyazit Yalcinkaya}\Email{beyazit@berkeley.edu}\\\addr University of California, Berkeley \AND
\Name{Lasse Peters}\Email{l.peters@tudelft.nl}\\\addr TU Delft \AND
\Name{David {Fridovich-Keil}}\Email{dfk@utexas.edu}\\\addr The University of Texas at Austin \AND
\Name{Georgios Bakirtzis}\Email{bakirtzis@telecom-paris.fr}\\\addr LTCI, Télécom Paris, Institut Polytechnique de Paris 
}

\usepackage{xurl}
\usepackage{mathtools}
\usepackage{xcolor}
\usepackage{zref-clever}
\zcRefTypeSetup{theorem}{
    Name-sg = Definition,
}
\usepackage{times}
\usepackage{transparent}
\usepackage{multirow,booktabs,graphicx,import}
\usepackage{array}
\usepackage{pgfplots}
\usepackage{amsmath,mathtools}
\pgfplotsset{compat=newest}
\usepackage{circledsteps}
\usepackage{enumitem}
\usepackage{doi}
\usepackage{hyperref}

\definecolor{lightred}{HTML}{ff0f19}

\newcommand{\blfootnote}[1]{%
  \begingroup
  \renewcommand\thefootnote{}%
  \footnote{#1}%
  \addtocounter{footnote}{-1}%
  \endgroup
}

\makeatletter
\renewcommand\@makefntext[1]{%
  \parindent 1em\noindent
  \hb@xt@1.8em{\hss\@makefnmark}#1%
}
\makeatother

\renewcommand{\vec}[1]{\mathbf{#1}}
\newcommand{\Nobjectives}{D}
\newcommand{\Aspace}{\mathcal{A}}
\newcommand{\Sspace}{\mathcal{S}}
\newcommand{\rewards}{\vec{r}}
\newcommand{\values}{\vec{v}}
\newcommand{\returns}{\vec{v}}
\newcommand{\return}{v}
\newcommand{\weights}{\vec{w}}
\newcommand{\dynamics}{P}
\newcommand{\RealsSpace}{\mathbb{R}}
\newcommand{\ObjectiveSpace}{\RealsSpace^\Nobjectives}
\newcommand{\Expectation}{\mathbb{E}}
\newcommand{\ParetoFrontier}{\mathcal{F}}

\date{\today}

\title[Controllability in preference-conditioned MORL]{Controllability in preference-conditioned\\multi-objective reinforcement learning}

\hypersetup{
    pdfauthor={Pau de las Heras Molins, Beyazit Yalcinkaya, Lasse Peters, David Fridovich-Keil, Georgios Bakirtzis},
    pdftitle={Controllability in preference-conditioned multi-objective reinforcement learning},
    pdfkeywords={},
    pdfsubject={},
    pdfcreator={},
    pdflang={English}
}

\hypersetup{colorlinks=true, citecolor=gray, linkcolor=gray, urlcolor=gray}

\begin{document}

\maketitle

\begin{abstract}
Multi-objective reinforcement learning (MORL) allows
a user to express preference over outcomes
in terms of the relative importance of the objectives,
but standard metrics cannot capture whether
changes in preference reliably change
the agent’s behavior in the intended way,
a property termed controllability.
As a result, preference-conditioned agents can score well
on standard MORL metrics while being insensitive to the preference input.
If the ability to control agents cannot be reliably assessed,
the symbolic interface that MORL provides
between user intent and agent behavior is broken.
Mainstream MORL metrics alone fail
to measure the controllability of preference-conditioned agents,
motivating a complementary metric specifically designed to that end.
We hope the results spur discussion in the community
on existing evaluation protocols to consolidate advances
in preference adaptation in MORL to larger and more complex problems.
\end{abstract}


\blfootnote{Code available at \href{https://codeberg.org/MustardWatches/PufferMO/}{https://codeberg.org/MustardWatches/PufferMO/}.}

\section{Introduction}

There are often many ways to solve the same task.
Which way is the ``best'', however,
often depends on \emph{preference}
for some options over others.
Instilling agents with the ability to find
multiple diverse solutions to problems
is key to generalization.
Often omitted is the equally important ability to adhere
to the preference over those diverse solutions,
such that the behavior aligns with what is desired.
Reinforcement learning (RL), in particular,
produces black-box behavior which
can be hard to understand and especially to \emph{control}.
Neurosymbolic algorithms address this by providing symbolic interfaces, such as task specifications or constraints \citep{yalcinkaya:2024,yalcinkaya:2025,jothimurugan:2025,vaezipoor:2021,jackermeier:2025}.
Symbolic interfaces augment learning with interpretable,
synthesizable and specifiable agent behavior.

Multi-objective RL (MORL) decomposes agent rewards
into distinct objectives describing
different aspects of task success \citep{hayes:2022}.
When rewards conflict, maximizing all of them
simultaneously might not be possible.
In a driving scenario, for instance, a reward for staying in lane
might conflict with a penalty for being too close to an obstacle ahead.
When there is no single ``best'' solution,
preference must be elicited
to choose one among the rest.

Prior work on multi-objective decision making has mainly studied preference
encoded as a ranking of the objectives by importance
\citep{wray:2015,zanardi:2022,lee:2025,delasheras:2025,rustagi:2025},
or as relative weights \citep{lin:2024,yang:2025,alegre:2023}.
A recurrent object of study in MORL is \emph{preference adaptation},
the ability of agents to adjust their behavior
to comply with a particular preference while solving a task.
Preference-conditioned policies in MORL expose a symbolic interface that allows users to \emph{control} solutions by modulating the relative importance of multiple human-understandable rewards.

Evaluating preference adaptation in MORL agents
is not without some challenges.
Most works report
a limited set of general metrics from the broader
multi-objective optimization field,
such as hypervolume and sparsity \citep{hayes:2022}.
These metrics, however, fail to capture \emph{controllability},
the relationship between changes in preference and induced changes in agent behavior.
Without reliable ways to measure agent controllability,
any assessment of preference adaptation is incomplete.
A MORL agent that completely ignores the user's preference might still
be deemed as ``good'' by mainstream metrics,
clearly showing a blind spot in current evaluation protocols.

Beyond the lack of metrics for controllability,
a practical obstacle for proper preference adaptation
evaluation in MORL is related to the field's toolkits.
MORL toolkits lack optimizations, limiting their throughput.
As a consequence, preference adaptation methods are often tested in a small number
of relatively simple environments and over limited training horizons,
making it difficult to assess how well they scale
or generalize to more complex problems.
The practical bottleneck has significant implications:
researchers may introduce increasingly complex algorithmic extensions
to compensate for limited experimental throughput,
further increasing computational costs and constraining the scope of empirical evaluation.

To make progress in these challenges, we:
\begin{enumerate}[nosep, noitemsep]
    \item characterize how standard MORL evaluation metrics can obscure failures of preference adaptation, motivating a more refined notion of \emph{controllability} and rank-correlation-based evaluation metrics; and
    \item publish PufferMO \citep{puffermo}, a MORL-compatible extension of PufferLib \citep{suarez:2024}, a high-throughput framework for RL in complex environments.
\end{enumerate}

\section{Related work}

\smallskip \noindent
\textsc{preference adaptation in morl} \quad
Research on RL agents that adapt to changing preferences
exists before MORL \citep{natarajan:2005}.
\citet{zanardi:2022}, \citet{rustagi:2025}, \citet{lee:2025} and \citet{delasheras:2025}
are examples of works that encode preference
as a strict ranking of the objectives.
A wider body of work has focused on
the non-preemptive trade-offs enabled by scalarization, for example by combining the objectives
using linear weights, often achieving dynamic adaptation
via preference-conditioned policies \citep{alegre:2023,alegre:2025}.
Most of these works report only metrics
used to evaluate solution sets in the broader multi-objective optimization field,
such as hypervolume and sparsity,
focusing on frontier quality while neglecting
the measurement of controllability.
Only a few works report metrics for preference adaptation
\citep{basaklar:2023,yang:2025,jiang:2026},
but even in these cases, the metrics are typically
coarsely aggregated across all objectives
and cannot always capture relevant aspects of controllability,
such as the directionality of changes in behavior
with respect to changes in preference.
Unlike previous work, our definition of controllability
explicitly states the assumptions about how conditioning preferences map to induced behavior,
revealing previously untapped insights.

\smallskip \noindent
\textsc{performance metrics in morl} \quad
Quantifying success in multi-objective problems is hard.
Naturally, no single metric is without limitation \citep{knowles:2002},
such as sensitivity to different scales between objectives
or requiring access to a reference solution set.
Combining multiple metrics is therefore essential
to capture different aspects of solution quality \citep{zitzler:2003}.
However, the sheer volume of multi-objective metrics \citep{audet:2021}
raises questions about which ones might be better suited for any particular setting---often chosen depending on the properties of interest, such as distribution,
convergence, or spread \citep{audet:2021}.
We posit that controllability is an essential property for studying the performance of multi-objective problems in general and present a focused definition and correlation-based
metrics to analyze controllability in the linear preference setting in particular.

\smallskip \noindent
\textsc{high-throughput toolkits in rl} \quad
Slow, low-throughput toolkits are a persistent problem in RL,
and MORL is especially susceptible: it is strictly harder than single-objective RL,
and existing MORL-capable toolkits have lagged behind broader advances in tools.
Practical limitations lead to long training times with reduced horizons,
incentivizing researchers to introduce increasingly complex algorithmic extensions
while still potentially causing reported improvements to fail to generalize
or be outperformed by hyperparameter tuning alone \citep{suarez:2024}.
PufferLib \citep{suarez:2025} is an RL library
designed for high-performance training in complex environments, such as autonomous driving  \citep{cornelisse:2025}.
In this work, we extend the functionality of PufferLib to the MORL setting,
enabling scalable experimentation on increasingly complex environments.

\section{Preliminaries}

We briefly define the problem formulation, structures, and preference-conditioning in MORL.

\subsection{MORL}

MORL models the environment as a
multi-objective Markov decision process (MOMDP) \citep{felten:2024}.
A MOMDP is composed of a state space $\Sspace$,
an action space $\Aspace$, dynamics
$\dynamics(s_{t + 1} \mid s_t, a_t)$,
a vector-valued reward function for $\Nobjectives$ objectives
$\rewards \colon \Sspace \times \Aspace \times \Sspace \to \RealsSpace^D$,
an initial state distribution $\mu(s_0)$,
and a discount factor $\gamma \in [0, 1)$,
packaged as the tuple $M \coloneq (\Sspace, \Aspace, \dynamics, \rewards, \mu, \gamma)$.

The solution space in a MOMDP consists
of the space of all possible policies $\Pi$.
Let a policy be defined as a mapping from states
to distributions over actions, $\pi(a \mid s)$.
We evaluate policies executed in a MOMDP $M$ using
the multi-objective \emph{action-value} and \emph{value} functions.

The multi-objective action-value function of a policy $\pi$,
$\vec{q} \colon \Sspace \times \Aspace \to \RealsSpace^D$,
is defined as
\[
    \vec{q}^\pi(s,a) \coloneq \Expectation_\pi \left[
        \sum_{i=0}^\infty \gamma^i \rewards_{t+i} \, \bigg|\, s_t=s, \, a_t=a
    \right],
\]
where $\Expectation_\pi$ denotes the expectation over trajectories induced by $\pi$.

The multi-objective value function of a policy $\pi$, $\values^\pi \colon \Sspace \to \RealsSpace^D$,
is defined as
\[
    \values^\pi(s) \coloneq \Expectation_\pi \left[\vec{q}^\pi(s,\pi(s))\right].
\]

The multi-objective advantage function of a policy $\pi$,
$\vec{a} \colon \Sspace \times \Aspace \to \RealsSpace^D$,
is defined as
\[\vec{a}^\pi(s,a) \coloneq \vec{q}^\pi(s,a) - \values^\pi(s).\]

Finally, multi-objective expected discounted return of a policy under initial state distribution $\mu$
is given by $\returns^\pi \coloneq \Expectation_{s_0 \sim \mu} \values^\pi(s_0)$.

In multi-objective problems, there is rarely a single solution
that achieves maximum returns $\returns^\pi$ on all objectives simultaneously.
Instead, there exists a set of incomparable returns known as the Pareto frontier.
Given a Pareto dominance relation $\succ_p$, the Pareto frontier of a MOMDP
is a set constructed from non-dominated returns $\returns^\pi$,
\[
\ParetoFrontier \coloneq \big\{ \returns^\pi \mid \nexists \pi'
\text{ such that } \returns^{\pi'} \succ_p \returns^\pi \big\}.
\]
A policy $\pi$ (Pareto-)dominates another policy $\pi'$
if it performs as well as $\pi'$ on all objectives
and does better in at least one of them,
\[
\returns^\pi \succ_p \returns^{\pi'}
\Longleftrightarrow
\big(\forall i : \return_i^\pi \geq \return_i^{\pi'}\big)
\wedge
\big(\exists i : \return_i^\pi > \return_i^{\pi'}\big).
\]

The space of stochastic policies
in a MOMDP is infinite,
rendering the search for all solutions
with returns in the Pareto frontier
an intractable problem.
Instead, the goal in MORL is typically
to learn policies whose returns approximate
those in the true Pareto frontier.
At inference time, the choice of the particular solution
depends on \emph{preference}.

\subsection{Linear preference in MORL}
\label{sec:preference_morl}

A common approach to describing preference in MORL
is as linear weights encoding the relative importance of objectives.
Augment a MOMDP $M$ with $\Nobjectives$ objectives
with the space of all potential preferences $\Delta^{\Nobjectives}$,
the $(\Nobjectives - 1)\text{-dimensional}$ simplex, such that
\[
\Delta^{\Nobjectives} \coloneq \biggl\{
    \weights \in \RealsSpace^{\Nobjectives} \biggm|
    \sum_{i=1}^{\Nobjectives} w_i = 1, \; w_i \ge 0
    \text{ for } i=1,\dots,\Nobjectives
\biggr\}.
\]
Each particular preference is represented by a weight vector $\weights \in \Delta^{\Nobjectives}$.
The scalar value $v^\pi$ of a policy $\pi$ considering preference $\weights$,
denoted as $v^\pi_\weights$, is given by the dot product $\cdot$ between vectors
\[
    v^\pi_\weights \coloneq \returns^\pi \cdot \weights.
\]

Linear scalarization cannot select solutions
in non-convex regions of the Pareto frontier \citep{lin:2024}
and cannot enforce strict priority ranking over the objectives
when they exist.
Despite these limitations, linear preference
is ubiquitous in enforcing preference in MORL, it is algorithmically straightforward, and provides interpretable weights
to understand relative objective importance.
For these reasons, this work focuses on linear preference,
although the discussion on controllability is orthogonal
to the preference representation approach.

Given two similar linear preferences,
the optimal policies for a task
often share some underlying structure.
On the one hand, training them separately, in this case,
reduces learning efficiency
by having to learn similar behaviors twice.
On the other hand, training multiple policies simultaneously
can be impractical due to compute constraints.

Training a single, preference-conditioned policy
alleviates these problems
\citep{abels:2019,basaklar:2023,alegre:2023,alegre:2025,yang:2025},
common also 
in goal-conditioned RL \citep{liu:2022}.
In the linear preference setting, preference-conditioned policies
are often parametrized by a weight vector.
During training, a new weight vector $\weights \sim \Delta^\Nobjectives$
is typically sampled for each episode \citep{alegre:2023,alegre:2025}.
At each training iteration,
the policy is updated based on a loss function
that uses weights $\weights$ to scalarize multi-objective values.
During inference, the weight-conditioned policy
$\pi(s; \weights)$
outputs a probability distribution over actions
that is expected to maximize utility according to $\weights$.
This effectively enables the agent to adapt to \emph{any} linear preference,
which might even change dynamically during an episode,
without requiring any retraining.

\subsection{Weight-conditioned multi-objective proximal policy optimization}
\label{sec:extending_ppo}

Proximal policy optimization (PPO)
is one of the most widely used RL algorithms
due to its training stability and efficiency \citep{schulman:2017}.
PPO has been extended to MORL with linear preference \citep{terekhov:2024},
in which case it is referred to as multi-objective PPO (MOPPO).
Both actor and critic networks in MOPPO
are conditioned on linear preference weights $\weights$.
Additionally, the critic head is augmented
with an additional dimension for the multiple objectives
(\zcref[S]{fig:architecture_extensions}).
During training with MOPPO, a critic network approximates
the multi-objective state value function as
\[
    \tilde{\values}^\pi(s; \weights)
    \coloneq \Expectation_\pi \left[ \sum_{t \geq 0} \gamma^t \tilde{\rewards}_t \biggm| s_0 = s \right],
\]
where $\tilde{\rewards}$ is the empirical reward obtained
by interacting with the environment.
The multi-objective advantage estimate $\tilde{\vec{a}}^\pi$ is obtained
through generalized advantage estimation \citep{schulman:2016}
using $\tilde{\values}^\pi(s; \weights)$.
Then, the policy term in the loss function used to update the networks
scalarizes $\tilde{\vec{a}}^\pi$ using the conditioning weights $\weights$ as
\[
    \Expectation_\pi \left[
        \sum_{t \geq 0} \, \log \pi(a_t; s_t) \; \tilde{\vec{a}}^\pi_t \cdot \weights
    \right],
\]
incentivizing the learning of a policy
that maximizes utility
under any conditioning preference.

\begin{figure}[!t]
    \centering
    \def\svgwidth{0.8\linewidth}
    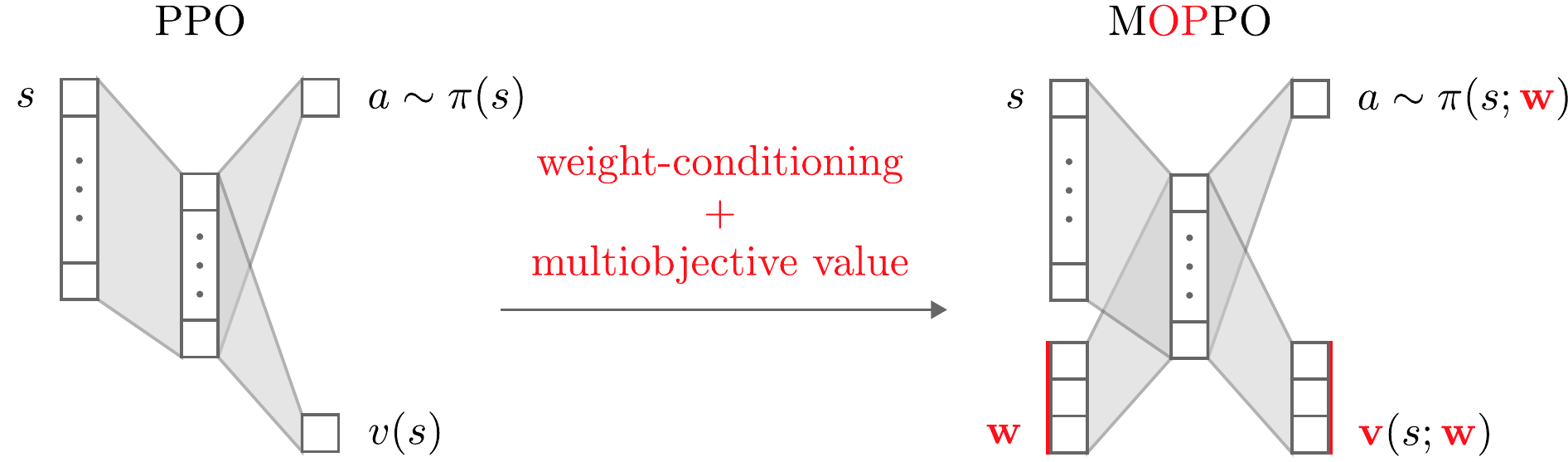
    \caption{
        MOPPO extends PPO with weight-conditioning and a multi-objective value head.
    }
    \label{fig:architecture_extensions}
\end{figure}

It is possible to train agents with
MOPPO without actually
conditioning the networks on the preference weights $\weights$.
This is equivalent to the marginalization of $\weights$,
yielding a multi-objective policy $\pi(s)$
that maximizes the expected multi-objective returns over the preference space.
In our experiments we use this non-conditioned variant of MOPPO
as a non-controllable baseline.

\section{Evaluating preference adaptation in MORL}

In this work, we show that mainstream metrics in MORL
are not enough to fully characterize preference adaptation in agents.
In this section, we briefly describe the most used
performance indicators in MORL \citep{hayes:2022}
along with their limitations (\zcref[S]{fig:metrics}).
We then present a new class of metrics based on rank correlation
that better capture \emph{controllability}.

The performance of MORL agents is usually analyzed in terms
of their attained solutions in the multi-objective space $\ObjectiveSpace$.
By rolling out policy $\pi$ for $K$ episodes in an environment,
we get a set of empirical returns $\{\returns^\pi\}$
that forms the agent's solution set.
We refer to the set of non-dominated returns in $\{\returns^\pi\}$
as the Pareto frontier approximation of the MORL agent,
denoted as $\ParetoFrontier$.

\subsection{Hypervolume}

Arguably the most commonly reported metric in MORL
is the hypervolume metric (HV) \citep{zitzler:1998}.
HV is the total hypervolume of the objective space $\ObjectiveSpace$
weakly dominated by $\tilde{\ParetoFrontier}$
and lower bounded by a reference point $\rewards \in \ObjectiveSpace$,
\[
\mathrm{HV}(\tilde{\ParetoFrontier}; \rewards) \coloneq \mathrm{Vol}\left(\bigcup_{\returns^\pi \in \tilde{\ParetoFrontier}}[\returns^\pi, \rewards]\right),
\]
where $[\returns^\pi, \rewards]$ is the $\Nobjectives$-dimensional hyperbox
$[\returns^\pi, \rewards] \coloneq \{\returns' \in \ObjectiveSpace \mid \rewards \leq \returns' \leq \returns^\pi\}$.

A set of solutions that completely dominates another always has greater HV.
HV is desirable for this strict monotonicity,
a property that many other multi-objective metrics lack \citep{audet:2021}.
However, the choice of reference point $\rewards$
and the normalization scheme (if any) of multi-objective returns
can have a large impact on the result \citep{wang:2023},
and the exact calculation of the HV can quickly become expensive
in high-dimensional objective spaces \citep{guerreiro:2022}.
Additionally, HV is biased towards central regions
of the objective space \citep{zitzler:2007},
where the volume increase with distance is largest (\zcref[S]{fig:metrics}).
HV serves as a proxy for evaluating solution diversity.
When it comes to preference adaptation, however,
HV fails to capture the extent to which each individual solution
complies with its inducing preference.

\subsection{Sparsity}

Sparsity (SP) \citep{xu:2020} is defined as the average distance
between solutions in the objective space.
Let $\tilde{\ParetoFrontier}^d$ be the ordered sequence of
elements in $\tilde{\ParetoFrontier}$ corresponding to objective dimension $d \leq \Nobjectives$,
and $\tilde{\ParetoFrontier}^d_i$ be the $i\text{th}$ element in the sequence.
SP is defined as the average distance between per-objective ordered returns in $\tilde{\ParetoFrontier}$,
\[
\mathrm{SP}(\tilde{\ParetoFrontier}) \coloneq
\frac{1}{|\tilde{\ParetoFrontier}|-1}\sum_{d=1}^{\Nobjectives}\sum_{i=1}^{|\tilde{\mathcal{F}}|-1}
\left(\tilde{\ParetoFrontier}^d_i-\tilde{\ParetoFrontier}^d_{i+1}\right)^2.
\]

The lower the SP, the denser the set of solutions,
potentially allowing a more fine-grained selection of desirable policies.
SP is susceptible to the differences in scale between objectives,
and can be misleading when comparing solution sets of varying cardinality.

\subsection{Expected utility}

Expected utility (EU) \citep{zintgraf:2015} measures the desirability
of a set of candidate solutions under a given preference model.
Unlike HV and SP, EU explicitly considers the existence of preference
in the form of utility functions.
For linear preference (\zcref[S]{sec:preference_morl}),
EU is defined as
\[
\mathrm{EU}(\tilde{\ParetoFrontier}) \coloneq \mathbb{E}_{\mathbf{w} \sim \Delta^\Nobjectives}\left[\max_{\mathbf{v}^\pi\in\tilde{\ParetoFrontier}}\mathbf{v}^\pi\cdot\mathbf{w}\right].
\]

If a set of solutions has high EU, it means that it \emph{could}
adapt to a wide range of preferences.
EU assumes that the agent is perfectly rational \citep{border:2020},
that is, the executed policy is the one
that maximizes utility under a given preference.
However, many algorithms typically cannot guarantee optimal policy selection,
as is the case in preference conditioning.
In such cases, the EU metric is misleading,
resulting in higher utility yields than are realistically achievable.

\begin{figure}[!t]
    \centering
    \def\svgwidth{1\linewidth}
    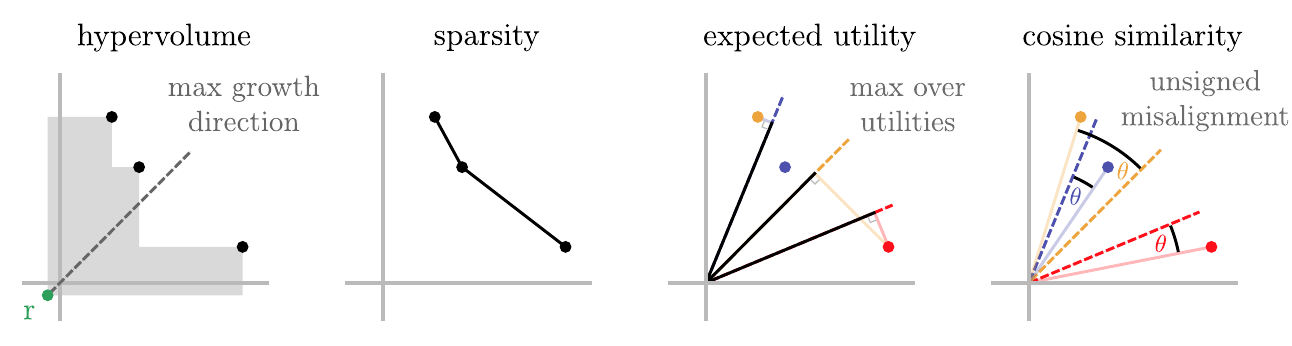
    \caption{
        Mainstream MORL metrics all have limitations.
        Dashed lines are linear preference weights,
        colored solutions are induced by corresponding preference.
        Hypervolume is biased towards inner regions.
        Expected utility assumes perfect rationality
        (max over utilities instead of actual $(\weights, \returns^\pi_\weights)$ pairs).
        Cosine similarity outputs unsigned misalignment.
    }
    \label{fig:metrics}
\end{figure}

\subsection{Measuring controllability}

Mainstream metrics for evaluating MORL algorithms
capture relevant aspects of the solutions attained by an agent.
However, they are not sufficient to capture all aspects
of successful preference adaptation.
Concretely, they fail to measure \emph{controllability},
the degree to which changes in preference yield the desired changes in behavior.
We present a formal definition of controllability
in MORL for linear preferences,
along with a corresponding family of metrics.

\begin{definition}[Controllability]
\label{def:controllability}
Let $\mathcal{M} \colon \Delta^{\Nobjectives} \to \RealsSpace^\Nobjectives$
denote a linear preference-conditioned agent's mapping from conditioning weights $\weights \in \Delta^\Nobjectives$
to induced multi-objective returns $\returns^\pi_\weights \in \RealsSpace^\Nobjectives$.
Let $f^d \colon \Delta^{\Nobjectives} \times \RealsSpace^\Nobjectives \to \RealsSpace$
be a function evaluating the quality of the mapping for objective $d$.
For each objective $1 \le d \le D$, define the controllability
with respect to objective $d$ as
\begin{equation}
\label{eq:controllability}
    \mathrm{CO}^d
    \coloneq \Expectation_{\weights \sim \Delta^{\Nobjectives}}
        f^d(\weights, \returns^{\pi}_\weights).
\end{equation}
An aggregate measure of controllability can then be defined as
\[
    \mathrm{CO}
    \coloneq \sum_{d=1}^{\Nobjectives} \mathrm{CO}^d.
\]
\end{definition}

\begin{remark}[Predictability and distinguishability]
Intuitively, controllability measures the extent to which
variations in the conditioning weights $\weights$
induce structured and distinguishable changes
in the resulting behavior. We say the mapping $\weights \mapsto \returns^\pi_\weights$
is:
(i) \emph{predictable} if it exhibits structured dependence
on $\weights$, and
(ii) \emph{distinguishable} if different preferences
induce measurably different returns. The choice of $f$ determines how these properties are quantified.
\end{remark}

We propose \emph{rank correlation} metrics
as the measure for mapping quality $f(\cdot)$
in \zcref[S]{def:controllability}, satisfying both stated properties.
The appropriate choice of rank correlation metric
depends on the assumptions made about controllability
for a particular preference representation scheme.
In the linear preference setting,
a natural (yet often implicit) assumption is that each objective's returns
vary monotonically with its corresponding inducing weight.
Kendall's rank correlation coefficient is appropriate
when monotonicity is the only assumption.
However, if the mapping is also expected to exhibit
some proportionality between weights and returns,
a better option is Spearman's
rank correlation coefficient $\rho$,
since it accounts for how large rank differences are.

An agent $\pi$ is conditioned on a set of $n$
linear preference vectors $W \coloneq \{\mathbf{w}_i\}_{i=1}^n$
and produces multi-objective returns $V \coloneq \{\mathbf{v}_i\}_{i=1}^n$.
For each objective $d \in \{1, \dots, \Nobjectives\}$,
let $W^d$ and $V^d$ be the sets of weights and returns obtained by extracting the $d\text{th}$ component from each element in $W$ and $V$, respectively.
The Spearman's correlation coefficient $\rho$ for objective $d$, denoted $\rho^d$, is defined as
\[
\rho^d(W, V) \coloneq
\frac{\mathrm{cov}\!\left(\mathrm{R}(W^d), \mathrm{R}(V^d)\right)}
    {\sigma_{\mathrm{R}(W^d)} \, \sigma_{\mathrm{R}(V^d)}},
\]
where $\mathrm{R}(\cdot)$ denotes the rank operator over $i \in \{1, \dots, n\}$.
The values of the coefficient $\rho^d \in [-1,1]$
range from $-1$, indicating inverse monotonicity,
to $1$, indicating strong monotonic alignment,
with $0$ indicating no ordinal association. Implementation details are provided in \zcref[S]{sec:metric_computation}.
Intuitively, an agent with high controllability in objective $d$
(as measured by $\rho^d$) will predictably attain
distinguishably higher returns for that objective
as its corresponding weight increases.

Recent work \citep{basaklar:2023, yang:2025}
has measured controllability for linear preference
using cosine similarity (CS) between
conditioning weights and induced returns,
acting as the mapping quality measure
$f^d(\cdot)$ in \zcref[S]{eq:controllability}
but without capturing the sign of the misalignment (\zcref[S]{fig:metrics}).
\citet{jiang:2026} use an ordering score
based on correlation between conditioning weights
and resulting returns.
These works, however, report only aggregate metrics
across all objectives, and do not provide insight
into how controllability differs for each of them.
Our definition of controllability is a generalization
of their approaches and is more explicit about the assumptions
underpinning the evaluation of preference adaptation.

\section{Experiments}
\label{sec:experiments}

The experiments consider a simple case
for controllability evaluation:
MOPPO, a controllable agent,
is compared against two baselines
that completely ignore preference,
PPO and MOPPO without weight conditioning (\zcref[S]{sec:extending_ppo}).
At a minimum, evaluation metrics ought to distinguish
any degree of controllability from none at all.
Yet the experiments show that mainstream MORL metrics
fail to consistently identify MOPPO
as the only controllable agent
and cannot offer insights on controllability when present,
while rank correlation provides
a clear measure for it.

The code is published as PufferMO \citep{puffermo}.
Experiments are carried out in multi-objective variants
of three complex environments: \textbf{Tetris}, \textbf{Snake}, and \textbf{MOBA}.
Further details on the environments are provided in \zcref[S]{sec:env_mod}.

We train 30 agents with random seeds
for each algorithm and environment.
Environments in PufferLib define a ``performance'' metric
quantifying task success beyond the returns
(game score in \textbf{Tetris}, average snake length,
win probability in \textbf{MOBA}).
We select the best runs based on the performance metric
and the discounted return, scalarized using uniform weights
for fair comparison with the single objective baseline.
Further details are provided in \zcref[S]{sec:parameters}.

\smallskip
\noindent
\textsc{evaluation scenario} \quad
During evaluation, we run 30 batches of parallel episodes
up to completion or termination and record the multi-objective returns.
For the MOPPO agent, each batch of episodes
uses one of 30 equally spaced weight vectors from the simplex.
We aggregate the discounted multi-objective returns
over batches and compute HV, SP, EU, CS and Spearman's per-objective coefficient $\rho^d$
(\zcref[S]{tab:evaluation}, see also \zcref[S]{sec:experimental_results}).
HV, SP and CS use min-max normalized returns
computed across algorithms within each environment.
The HV reference point is set $\mathbf{0.1}$
beyond the nadir point, following prior work \citep{wang:2023}.
EU uses the evaluation weights uniformly sampled over the simplex.

\begin{table}[!t]
\centering
\small
\makebox[\linewidth][c]{%
    \begin{tabular}{@{}ll*{2}{c}*{2}{c}*{3}{c}@{}}
    \toprule
    Env                & Algorithm         & HV $\uparrow$    & SP $\downarrow$   & EU $\uparrow$                 & CS $\uparrow$                & \multicolumn{3}{c}{Spearman $\rho^d$ $\uparrow$} \\
    \midrule
                    &                  &                 &                &                             &                            & $r_\text{death}$            & $r_\text{xp}$               & $r_\text{tower}$ \\ \cmidrule(l){7-9}
    \multirow{3}{*}{MOBA} 
                    & PPO              & \textbf{1.3302} & \textbf{0.4}            & \textbf{2.307 $\pm$ 1.243}  & 0.788 $\pm$ 0.115          & --                          & --                          & -- \\
                    & MOPPO (no cond.) & 0.3447          & 0.6            & 1.648 $\pm$ 0.858           & 0.764 $\pm$ 0.151          & --                          & --                          & -- \\
                    & \textcolor{lightred}{MOPPO} & 0.3651          & 7.2   & 1.690 $\pm$ 0.913           & \textbf{0.848 $\pm$ 0.109} & 0.520*                      & 0.737*                      & 0.056 \\
    \midrule
                    &                  &                 &                &                             &                            & $r_\text{food}$             & $r_\text{corpse}$           & $r_\text{death}$ \\ \cmidrule(l){7-9}
    \multirow{3}{*}{Snake} 
                    & PPO              & \textbf{1.1680} & 0.5            & \textbf{1.990 $\pm$ 1.147}  & \textbf{0.783 $\pm$ 0.121} & --                          & --                          & -- \\
                    & MOPPO (no cond.) & 0.3098          & \textbf{0.4}            & 1.762 $\pm$ 1.025           & 0.760 $\pm$ 0.160          & --                          & --                          & -- \\
                    & \textcolor{lightred}{MOPPO} & 1.0214          & 8.8   & 1.971 $\pm$ 1.129           & \textbf{0.783 $\pm$ 0.165} & -0.160                      & -0.655*                     & 0.959* \\
    \midrule
                    &                  &                 &                &                             &                            & $r_\text{combo}$            & $r_\text{drop}$             & $r_\text{rotate}$ \\ \cmidrule(l){7-9}
    \multirow{3}{*}{Tetris} 
                    & PPO              & 0.1268          & 0.1            & 5.680 $\pm$ 3.852           & 0.651 $\pm$ 0.260          & --                          & --                          & -- \\
                    & MOPPO (no cond.) & 0.1181          & \textbf{0.0}            & 5.706 $\pm$ 3.898           & 0.646 $\pm$ 0.263          & --                          & --                          & -- \\
                    & \textcolor{lightred}{MOPPO} & \textbf{0.4284} & 23.4  & \textbf{5.817 $\pm$ 3.787}  & \textbf{0.716 $\pm$ 0.212} & 0.874*                      & 0.733*                      & 0.695* \\
    \bottomrule
    \end{tabular}%
}
\caption{
    Evaluation metrics on the solutions by each algorithm.
    In bold, the ``best'' result per environment
    (e.g., $\uparrow$ means ``higher is generally better'').
    Mainstream metrics fail to consistently identify
    the only controllable algorithm, in red.
    Sparsity (SP) reported in units of $10^{-3}$. Asterisks denote significance $p < .001$.
}
\label{tab:evaluation}
\end{table}

\smallskip \noindent
\textsc{main result} \Circled{\oldstylenums 1} -- \textsc{mainstream morl metrics fail to measure controllability} \\
HV and EU do not consistently distinguish MOPPO from
the non-conditioned algorithms across all environments.
CS only achieves slightly higher values for MOPPO in both
\textbf{MOBA} and \textbf{Tetris}.
The difference does not reflect the fact
that only MOPPO has any controllability at all.
SP does clearly indicate a higher value for MOPPO across all environments.
While a typically undesirable result (a denser set is preferred),
it does identify MOPPO as the only weight-conditioned algorithm
through the spread of its solutions.
Still, it cannot measure whether solutions
comply with the given preference.
Mainstream metrics fail to quantify agent controllability
by focusing on aspects of the frontier's geometry,
neglecting the measurement of controllability
to the point of not always being able to distinguish
a controllable agent from others that cannot be controlled at all.

\begin{figure}[!t]
    \centering
    \includegraphics[width=\linewidth]{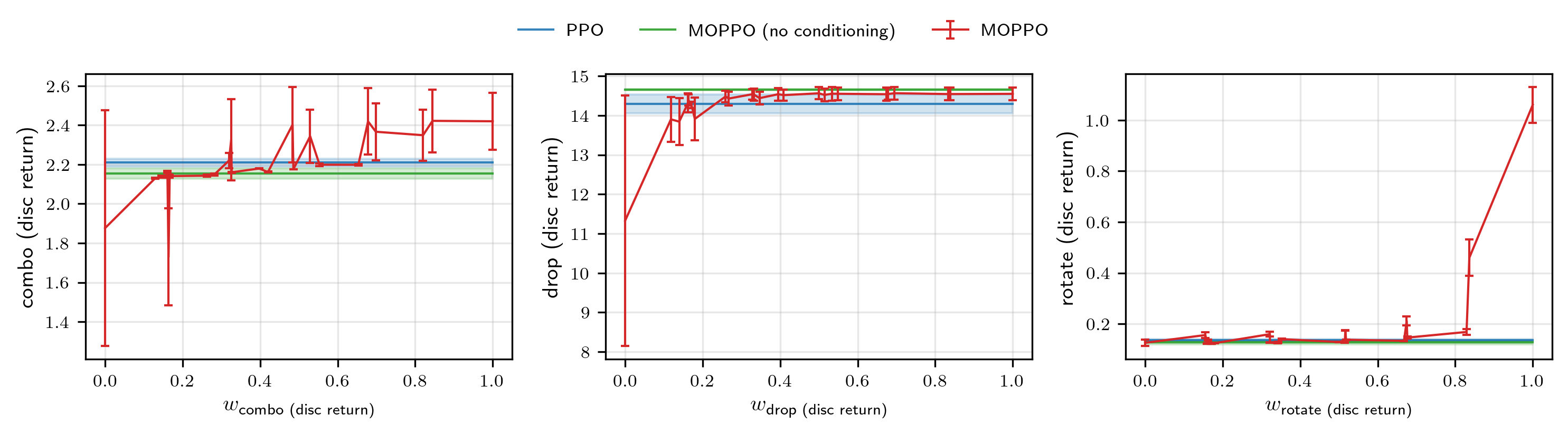}
    \caption{
        Per-objective discounted returns in \textbf{Tetris}.
        As the conditioning weight for an objective increases,
        so does the corresponding induced return by MOPPO,
        showing positive correlation.
        Shaded bands represent the mean $\pm$ one standard deviation
        of returns for non-conditioned algorithms (PPO and MOPPO without conditioning).
    }
    \label{fig:correlation_tetris}
\end{figure}

\smallskip \noindent
\textsc{main result} \Circled{\oldstylenums 2} -- \textsc{rank correlation offers insight into agent controllability} \\
Non-controllable algorithms
do not use preference during inference,
so their outputs are independent of evaluation weights,
making the computation of $\rho$ uninformative.
Still, the lack of solution diversity of non-controllable algorithms
can be assessed by the low variance in their returns
(\zcref[S]{fig:correlation_tetris}).
For the preference-conditioned agent, however,
$\rho$ becomes a direct and meaningful measure of controllability
by quantifying the relationship
between per-objective preference weights and induced returns
(\zcref[S]{fig:correlation_tetris} and \zcref[S]{tab:evaluation}).
A strong positive correlation,
such as the one observed in \textbf{Tetris},
indicates that MOPPO can successfully modulate its behavior
to produce results that comply with preference weights.
In other environments the correlation is not as strong or even negative (\zcref[S]{sec:experimental_results}),
signaling low controllability potentially due to problems
in reward design, the training process or structural limitations of the environment.
Unlike other metrics, rank correlation
offers insights into an agent's controllability,
filling an existing gap
in the evaluation of preference adaptation in MORL.

\section{Conclusion}

Preference adaptation in MORL
promises controllable agents,
but metrics typically used during evaluation
cannot adequately capture this property,
rendering any analysis potentially incomplete.
Our experiments show that rank-correlation-based controllability metrics are a valuable complement to existing ones by providing a missing perspective:
by explicitly analyzing the relationship between linear preferences and induced behavior,
they reveal to what extent agents are truly controllable.

This finding is a call to rethink how we validate preference-conditioned systems.
If we cannot reliably assess our ability to control agents,
the symbolic interface between human intent and agent behavior is broken.
Techniques for preference adaptation in MORL are
increasingly informing research on reward shaping,
agent alignment, and related areas.
Rigorous controllability evaluation is thus essential
to consolidate advances in the field.
In this work, we present a new metric
and a performant framework for proper controllability evaluation
of preference-conditioned agents
in complex environments.
The path forward requires not only more adaptable agents
even in challenging environments,
but metrics that hold them accountable to the preferences they claim to follow.

\acks{%
P. de las Heras Molins and G. Bakirtzis are partially supported by the academic and research chair
\emph{Architecture des Systèmes Complexes} through the following partners: Dassault Aviation, Naval
Group, Dassault Systèmes, KNDS France, Agence de l'Innovation de Défense, and
Institut Polytechnique de Paris.
D. Fridovich-Keil is partially supported by the U.S. National
Science Foundation~(NSF) under Grant No. 2336840.
}

\bibliography{manuscript}

\newpage
\appendix

\section{Additional experimental results}
\label{sec:experimental_results}

\subsection{Visualization of quantitative metrics}

The Pareto frontier approximations attained by the different algorithms
show that PPO tends to outperform both variants of MOPPO
(\zcref[S]{fig:pareto_fronts}).
A potential benefit of preference adaptation,
the flexibility to adapt to changing requirements,
is exemplified by the increased spread and diversity
of the solutions attained by MOPPO compared to the baselines
(even if at the cost of some performance in some cases).

Mainstream MORL metrics fail to consistently identify
the only controllable agent across environments
(\zcref[S]{fig:hypervolume_sparsity,fig:utility_cosine}),
with the exception of sparsity,
which identifies the greater spread of MOPPO's frontier
but does not say anything about how controllable the agent is.

Not all objectives are equally controllable.
For instance, the \texttt{corpse} objective in \textbf{Snake}
exhibits an inverted weight--return relationship,
hinting at reward design issues,
and in other environments some objectives plateau past a certain weight value
(\zcref[S]{fig:correlations}).
Rank correlation goes beyond flagging controllable agents:
the per-objective $\rho^d$ quantifies which objectives the agent
has better learned to trade off,
an insight unavailable from any mainstream metric (\zcref[S]{fig:spearman}).

\begin{figure}
    \centering
    \begin{minipage}[b]{\linewidth}
        \centering
        \includegraphics[width=\linewidth]{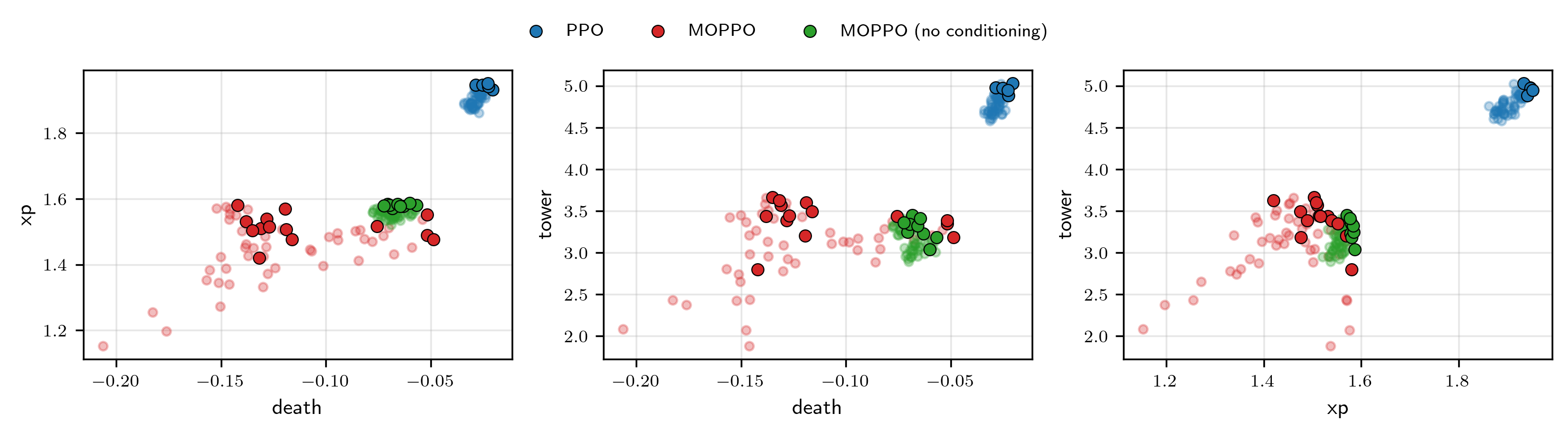}
        \par (a) MOBA
        \label{fig:pareto_moba}
    \end{minipage}
    \par\vspace{1ex}
    \begin{minipage}[b]{\linewidth}
        \centering
        \includegraphics[width=\linewidth]{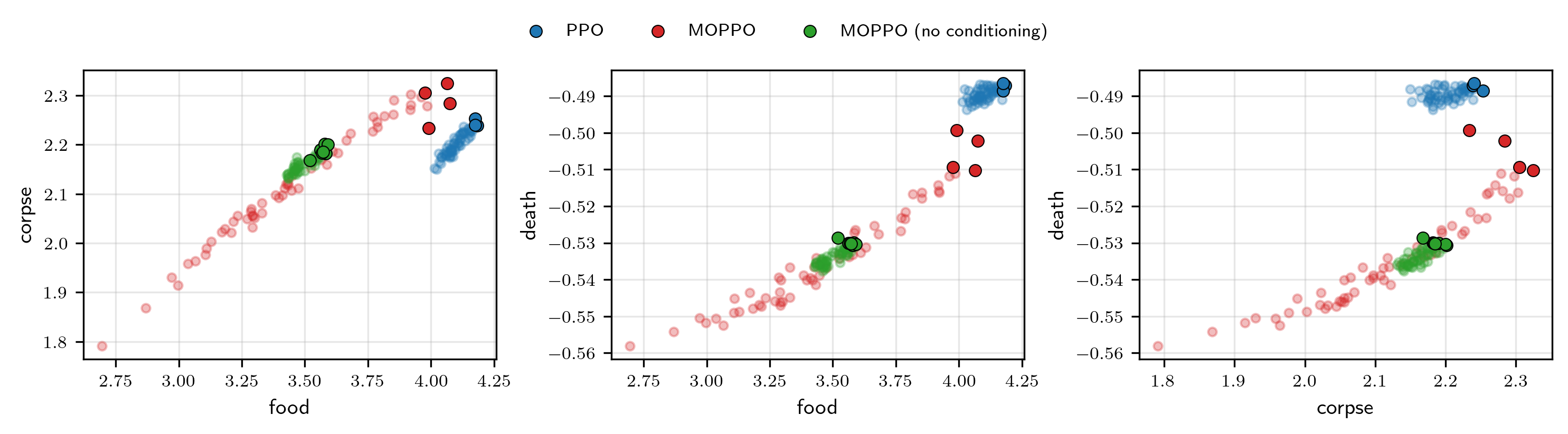}
        \par (b) Snake
        \label{fig:pareto_snake}
    \end{minipage}
    \par\vspace{1ex}
    \begin{minipage}[b]{\linewidth}
        \centering
        \includegraphics[width=\linewidth]{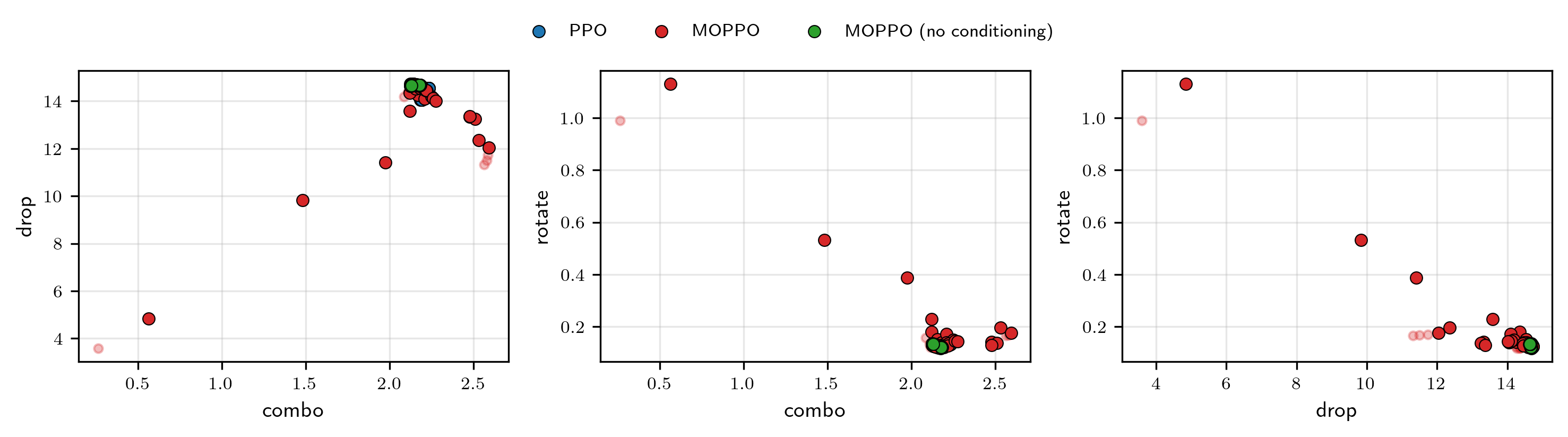}
        \par (c) Tetris
        \label{fig:pareto_tetris}
    \end{minipage}
    \caption{
        Preference conditioning trades peak performance for solution diversity.
        Each point is the average return of a batch of episodes;
        solid-bordered points are globally non-dominated.
        While PPO reaches higher-quality solutions, MOPPO's conditioning
        yields a wider spread across the objective space.
    }
    \label{fig:pareto_fronts}
\end{figure}

\begin{figure}
    \centering
    \begin{minipage}[t]{0.48\linewidth}
        \centering
        \includegraphics[width=\linewidth]{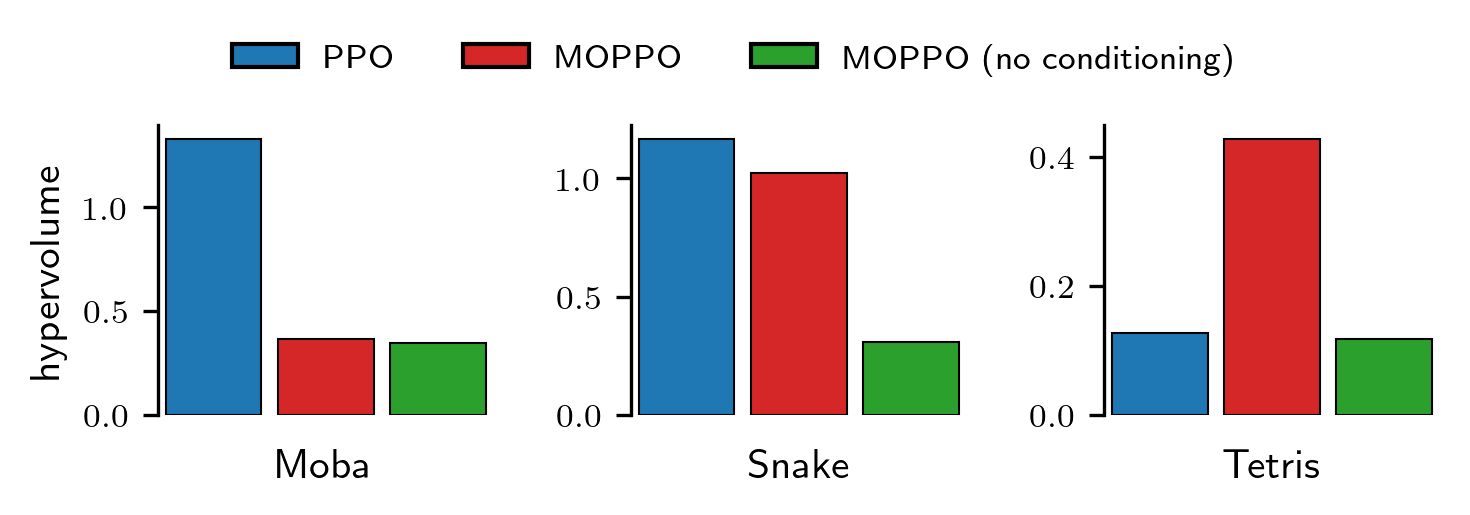}
        \par (a) Hypervolume
    \end{minipage}
    \hfill
    \begin{minipage}[t]{0.48\linewidth}
        \centering
        \includegraphics[width=\linewidth]{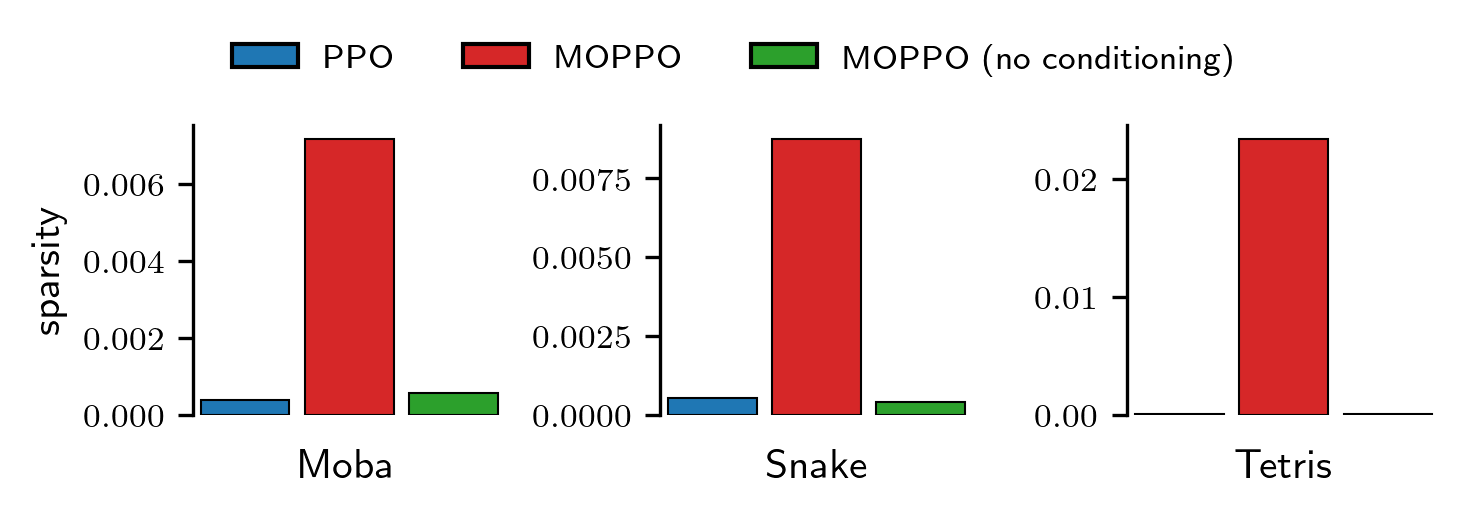}
        \par (b) Sparsity
    \end{minipage}
    \caption{
        Hypervolume and sparsity do not consistently identify MOPPO
        as the only controllable algorithm.
        (a) Hypervolume ($\uparrow$ better) is dominated by PPO in most environments.
        (b) Sparsity ($\downarrow$ better) does flag MOPPO's wider spread,
        but cannot assess whether individual solutions comply with their inducing preference.
    }
    \label{fig:hypervolume_sparsity}
\end{figure}

\begin{figure}
    \centering
    \begin{minipage}[t]{0.48\linewidth}
        \centering
        \includegraphics[width=\linewidth]{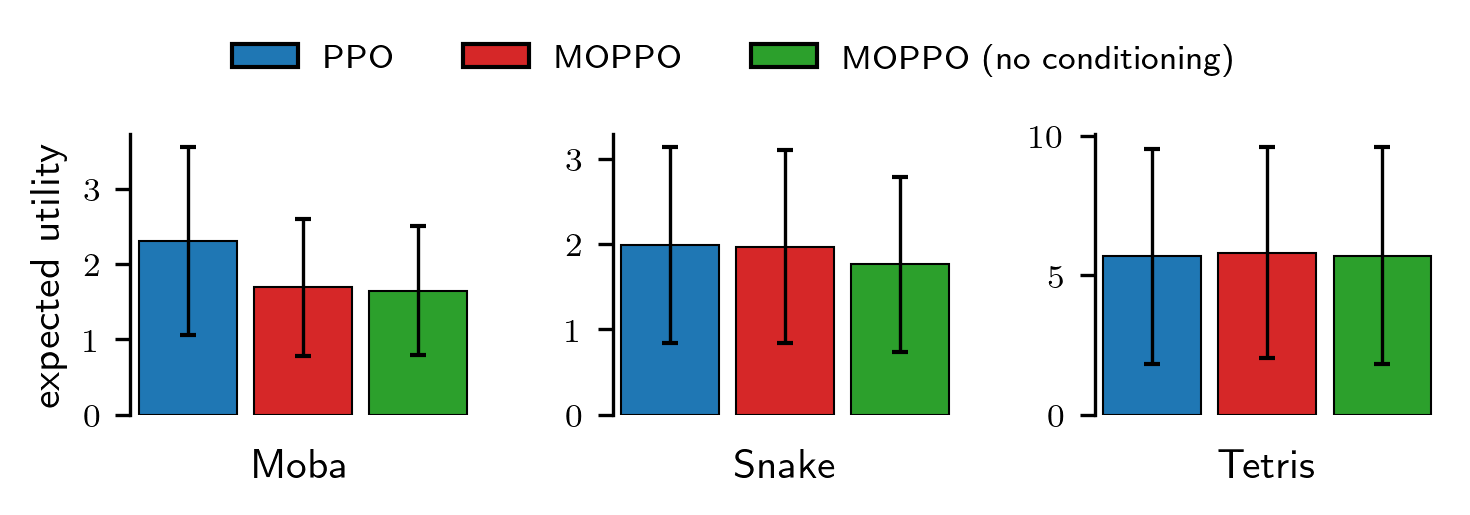}
        \par (a) Expected Utility
    \end{minipage}
    \hfill
    \begin{minipage}[t]{0.48\linewidth}
        \centering
        \includegraphics[width=\linewidth]{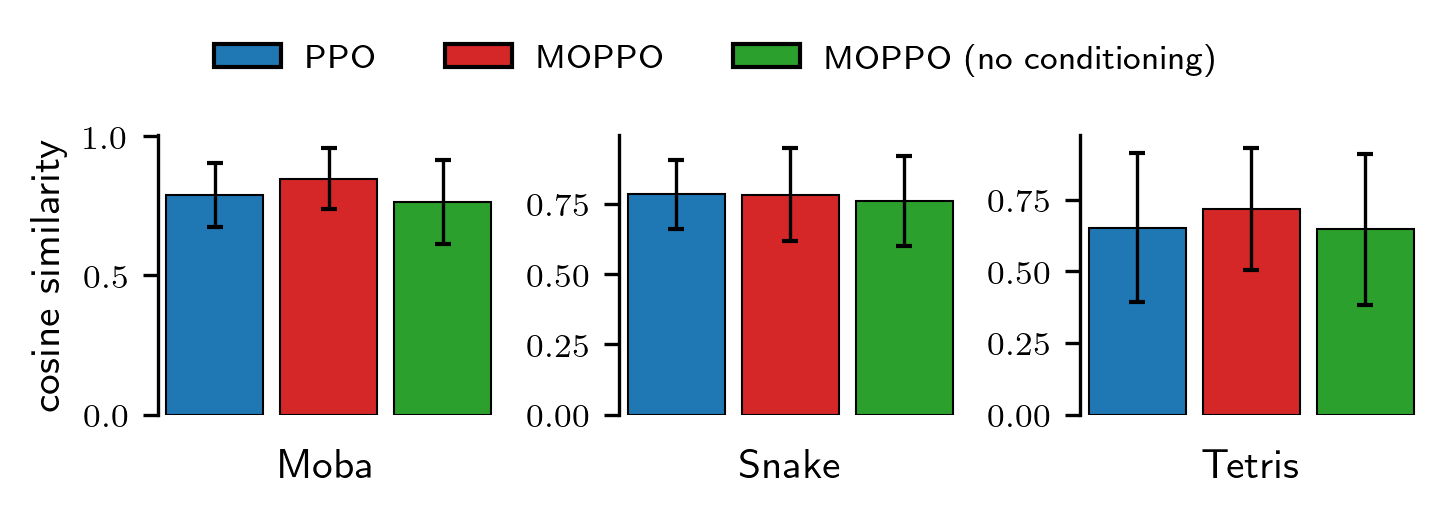}
        \par (b) Cosine Similarity
    \end{minipage}
    \caption{
        Expected utility and cosine similarity also fail to reliably flag MOPPO as the only controllable agent.
        (a) Expected utility ($\uparrow$ better) reflects overall solution quality,
        not whether each solution complies with its inducing preference.
        (b) Cosine similarity ($\uparrow$ better) shows only marginal differences between algorithms.
    }
    \label{fig:utility_cosine}
\end{figure}

\subsection{Qualitative evaluation of dynamic preference adaptation}
\label{sec:dynamic_adaptation}

We qualitatively assess how weight-conditioned policies
adapt to changing preferences at inference time.
An illustrative example in the \textbf{Tetris} environment
is showcased for its easily observable behavioral changes (\zcref[S]{fig:dynamic_adaptation}).

A MOPPO agent is run for a single episode of 1000 time steps (approximately 1 minute of gameplay).
The conditioning preference changes mid-episode (at time step 500),
with weights going from $\weights = [0.5, 0.5, 0.0]$ (that is, equal importance
to dropping pieces quickly and performing combos while ignoring rotation),
to $\weights = [0.0, 0.0, 1.0]$ (that is, only maximizing rotation).

We qualitatively observe how the agent adapts its behavior
to the new preference, noticeably rotating pieces
much more often after the change.
A small delay is observed between the preference change
and the shift in the per-objective rewards' trend
due to the inertia of an LSTM layer in the policy network.
This example demonstrates the potential for weight-conditioned policies
to dynamically adapt to changing preferences at inference time,
without requiring any additional training on the new preference.

\begin{figure}
    \centering
    \includegraphics[width=0.48\linewidth]{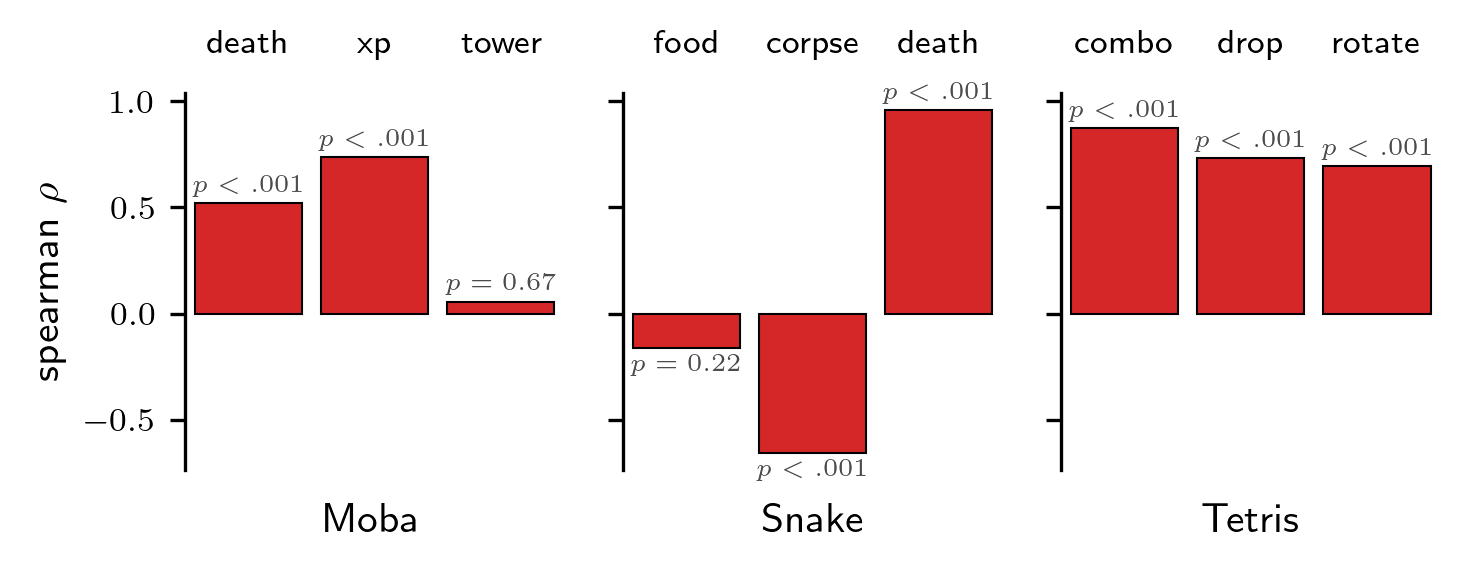}
    \caption{
        Rank correlation uniquely characterizes
        MOPPO's controllability.
        The per-objective breakdown exposes
        which objectives the agent has better learned to trade off,
        an insight unavailable from any mainstream metric.
    }
    \label{fig:spearman}
\end{figure}

\begin{figure}
    \centering
    \begin{minipage}[b]{0.9\linewidth}
        \centering
        \includegraphics[width=\linewidth]{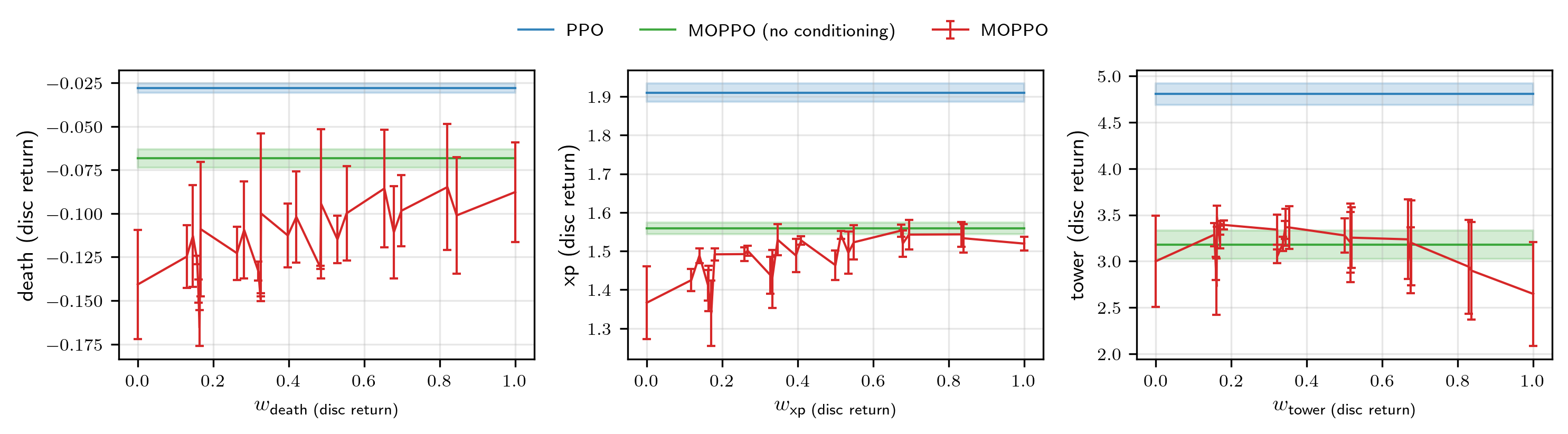}
        \par (a) MOBA
        \label{fig:correlations_moba}
    \end{minipage}
    \par\vspace{1ex}
    \begin{minipage}[b]{0.9\linewidth}
        \centering
        \includegraphics[width=\linewidth]{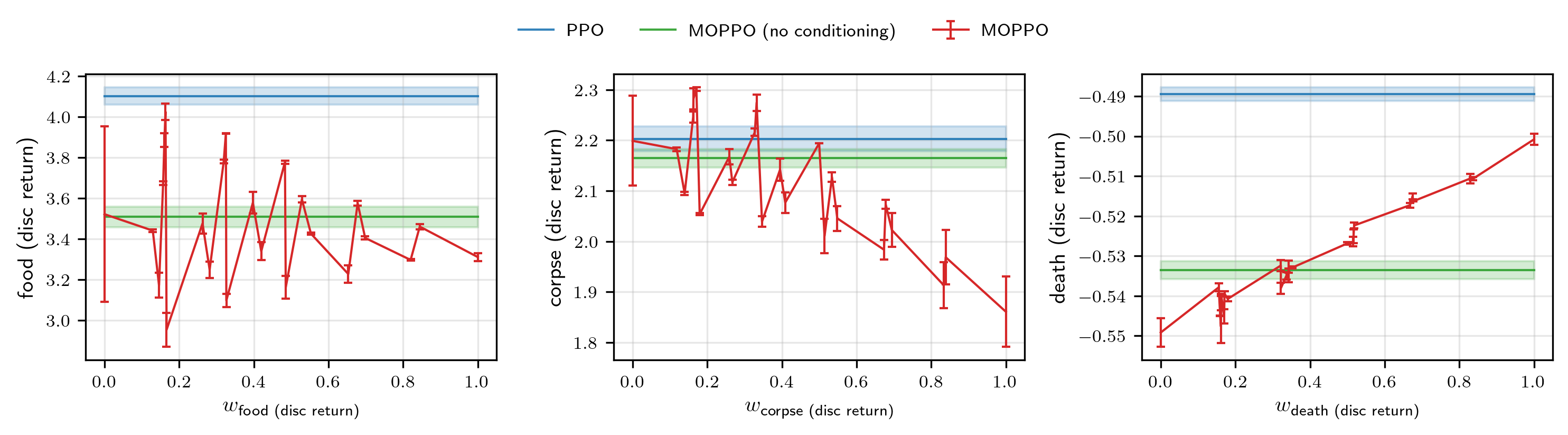}
        \par (b) Snake
        \label{fig:correlations_snake}
    \end{minipage}
    \par\vspace{1ex}
    \begin{minipage}[b]{0.9\linewidth}
        \centering
        \includegraphics[width=\linewidth]{figures/static_adaptation/correlation_scatter_tetris.png}
        \par (c) Tetris
        \label{fig:correlations_tetris}
    \end{minipage}
    \caption{
        Controllability varies heterogeneously
        across objectives and environments,
        revealing structural limits of preference adaptation.
        The \texttt{corpse} objective in \textbf{Snake}
        shows an inverted relationship,
        hinting at reward design issues.
        Shaded bands show the negligible variance
        in returns of non-conditioned baselines.
    }
    \label{fig:correlations}
\end{figure}

\begin{figure}
    \centering
    \begin{minipage}[t]{0.66\textwidth}
        \centering
        \def\svgwidth{\linewidth}
        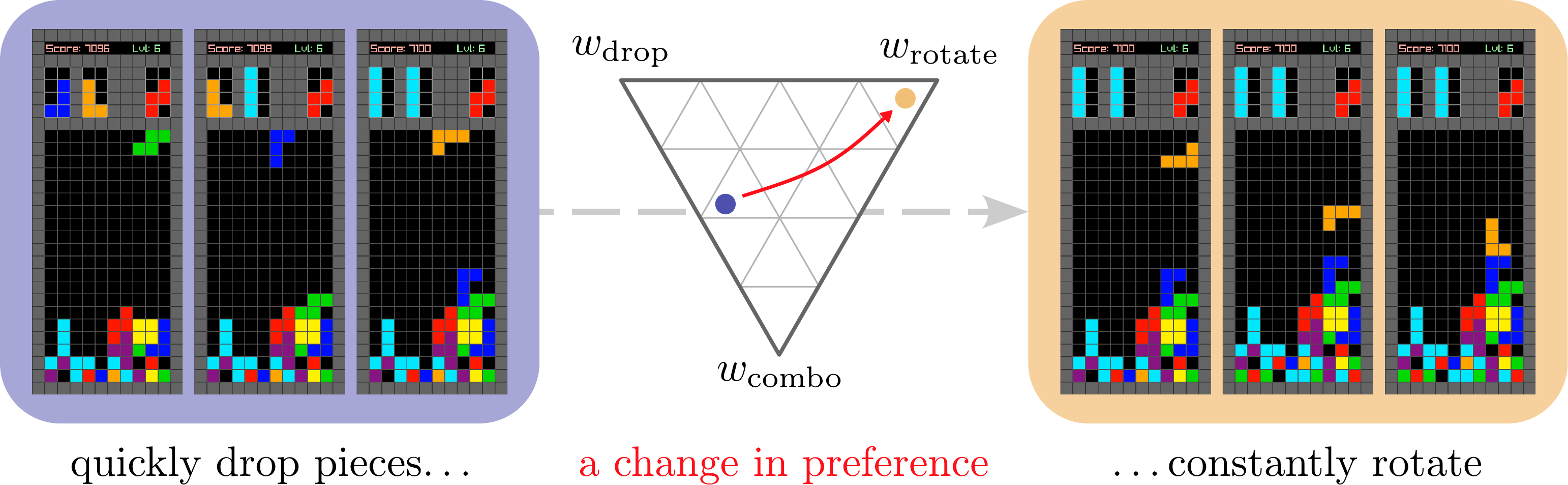
        \\ (a) Behavioral adaptation
        \label{fig:dynamic_adaptation_frames}
    \end{minipage}
    \hfill
    \begin{minipage}[t]{0.3\textwidth}
        \centering
        \includegraphics[width=\linewidth]{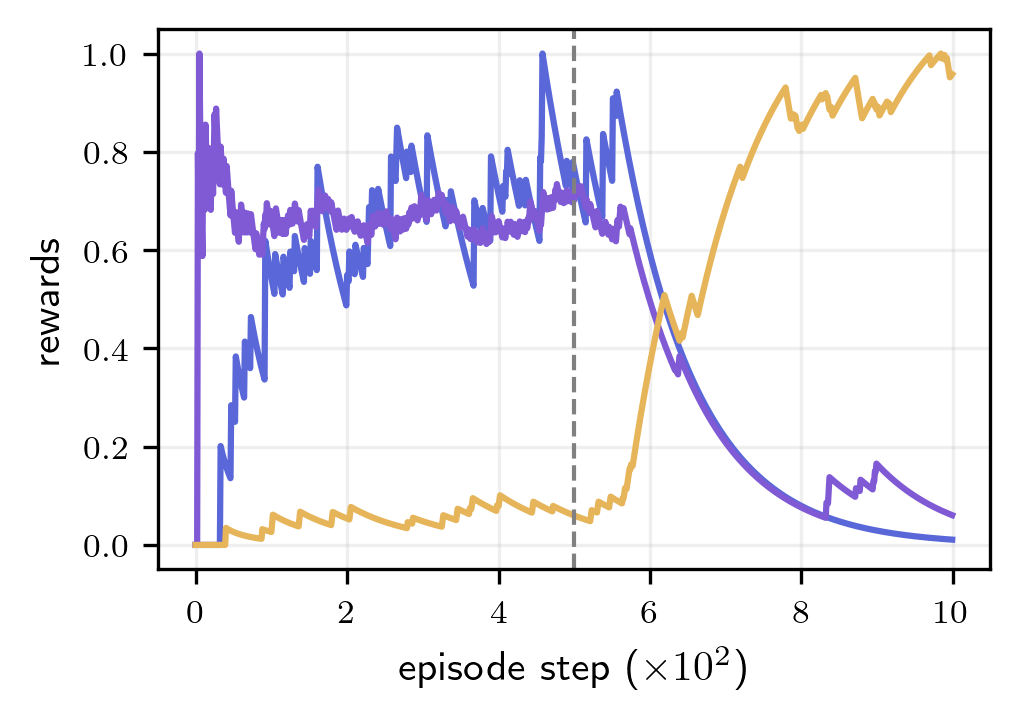}
        \\ (b) Rewards
        \label{fig:dynamic_adaptation_rewards}
    \end{minipage}
    \caption{
        Preference-conditioned MOPPO successfully adapts its behavior mid-episode without any retraining.
        (a) After the preference switches at step 500, the agent visibly rotates pieces more frequently.
        (b) A brief lag in the reward response reflects the inertia of the LSTM layer in the policy network.
    }
    \label{fig:dynamic_adaptation}
\end{figure}

\newpage

\section{Extending PufferLib to the multi-objective setting}

This section provides some technical details
on the modifications made to
\href{https://github.com/PufferAI/PufferLib/tree/3.0}{PufferLib 3.0} \citep{suarez:2024}
to support multi-objective environments and preference-conditioned policies.
All implementation details are available in the PufferMO repository \citep{puffermo}.
PufferLib 3.0 was extended to the multi-objective setting
through modifications spanning four architectural layers:
the environment binding interface and the vectorization layer,
the C/CUDA compute kernels, the training loop, and the policy models.

\smallskip \noindent
\textsc{environment and vectorization layer} \quad
A new C binding header was introduced alongside the existing interface
for multi-objective environments (\zcref[S]{sec:env_mod}),
adding a shared-memory slot for a per-step reward weight vector
alongside the vectorial reward buffer.
The serialization layer was extended to detect whether a wrapped environment
exposes a \texttt{reward\_dim} attribute;
when present, the environment returns
an additional weight tensor
alongside the standard information.
A new method was implemented to propagate
the weight vector through the vectorized environment stack
down to each individual environment instance,
enabling outer loops to broadcast the current scalarization weight vector at runtime.

\smallskip \noindent
\textsc{training loop} \quad
The rollout engine allocates per-step weight buffers
alongside the vectorial reward buffer when \texttt{reward\_dim > 1},
and expands the value buffer to match the additional objective dimension.
The rollout collection loop stores the weight tensors in the experience buffer.
Advantage estimation is handled by a new function
mirroring the existing scalar variant,
operating on rank-3 value and reward tensors
and dispatching to a dedicated CUDA kernel with a CPU fallback.
During the PPO update phase, scalarization is applied explicitly:
per-objective advantages are contracted with the stored weight vectors
to produce a scalar advantage for the policy gradient loss,
and the per-objective critic loss is similarly weighted before summing.

\smallskip \noindent
\textsc{policy models} \quad
The model classes used in the extended environments accept a new option to append the weight vector to the observation
before the encoder, enabling conditioned policies.
The recurrent wrapper propagates this flag
and handles the additional tensor reshape cases
for back-propagation-through-time unrolled value tensors.

\subsection{Creating multi-objective variants of PufferLib environments}
\label{sec:env_mod}

We selected three diverse environments from PufferLib to evaluate our multi-objective extensions:
\texttt{puffer\_moba}, \texttt{puffer\_tetris}, and \texttt{puffer\_snake} (\zcref[S]{fig:env_screenshots}).
These environments were chosen to represent different challenges:
\begin{itemize}
    \item \textbf{Tetris}: PufferLib's version of the classic single-agent game,
        where the player is rewarded for taking the hard drop and rotation actions
        as well as for clearing multiple lines at once.
        The difficulty of the game incrementally increases
        by having the pieces fall faster as the game progresses.
        \textbf{Fully observable, single-agent environment.}
    \item \textbf{Snake}: multi-agent version of the classic game,
        where multiple snakes compete for food in a shared grid world.
        Agents are rewarded for eating either food or other snakes' remains,
        and penalized for colliding with other snakes or the walls.
        \textbf{Fully observable, multi-agent environment.}    
    \item \textbf{MOBA}: simplified version of a multiplayer online
        battle arena game (for example, Dota 2, \citet{openai:2019}),
        where two teams of five agents compete to destroy the opposing team's base.
        Agents are rewarded for gathering experience points and destroying enemy structures, and penalized for dying.
        \textbf{Complex, high-dimensional, partially observable multi-agent environment.}
\end{itemize}

These environments lend themselves to the decomposition
of their rewards into distinct intuitive components
without synthetic modifications.
We created multi-objective variants by exposing individual reward components
as vectors instead of summing them. We maintained original reward magnitudes
(\zcref[S]{tab:reward_magnitudes}) to facilitate hyperparameter transfer,
except for the MOBA environment where we enabled a death penalty to encourage survival
(disabled by default in the original).

We enabled logging of per-objective returns in both
single and multi-objective environment variants
using the same discount factor as in training.
In the Snake environment, we added automatic episode truncation
with random jitter to enable periodic weight resampling
during MOPPO training.

\begin{figure}
    \centering
    \begin{minipage}[c]{0.58\linewidth}
        \centering
        \includegraphics[width=\linewidth]{figures/env\_screenshots/moba.png}
        \par \texttt{puffer\_moba}
        \par\vspace{1em}
        \includegraphics[width=\linewidth]{figures/env\_screenshots/snake.png}
        \par \texttt{puffer\_snake}
    \end{minipage}
    \hfill
    \begin{minipage}[c]{0.37\linewidth}
        \centering
        \includegraphics[width=0.8\linewidth]{figures/env\_screenshots/tetris.png}
        \par \texttt{puffer\_tetris}
    \end{minipage}
    \caption{\raggedright
        Three environments of diverse complexity serve as a testbed for evaluating controllability.
        \textbf{MOBA}: high-dimensional, partially observable multi-agent battle arena.
        \textbf{Snake}: competitive multi-agent grid world.
        \textbf{Tetris}: fully observable single-agent puzzle game.
    }    
    \label{fig:env_screenshots}
\end{figure}

\begin{table}
\renewcommand{\arraystretch}{1.3}
\setlength{\tabcolsep}{6pt}
\centering
\begin{tabular}{rlll}
\textbf{Environment} & \textbf{Reward} & \textbf{Magnitude} & \textbf{Description} \\
MOBA   & $r_\text{death}$    & $-1.0$                & Penalty for dying (only in MO variant)          \\
        & $r_\text{xp}$       & $1.69 \times 10^{-3}$ & Gain experience (e.g., kill enemies, minions, etc.) \\
        & $r_\text{tower}$    & $4.53$                & Destroy an enemy tower (shared by all players)  \\
Snake  & $r_\text{food}$     & $0.1$                 & Eat food                                        \\
        & $r_\text{corpse}$   & $0.01$                & Eat corpse of other snakes                      \\
        & $r_\text{death}$    & $-1.0$                & Penalty for colliding with walls or snakes      \\
Tetris & $r_\text{drop}$     & $0.02$                & Drop pieces quickly ("hard drop")               \\
        & $r_\text{combo}$    & $[0, 1]$              & Clear multiple lines ($\times$ number of lines) \\
        & $r_\text{rotate}$   & $0.01$                & Rotate pieces                                   \\
\end{tabular}
\caption{Reward magnitudes span several orders of magnitude across objectives, motivating normalization during evaluation.}
\label{tab:reward_magnitudes}
\end{table}

\section{Training configuration and experimental settings}
\label{sec:parameters}

PufferLib provides a default hyperparameter configuration file
along with environment-specific overrides.
While both \textbf{MOBA} and \textbf{Snake} environments
use the same default configuration for training,
\textbf{Tetris} has an environment-specific set of hyperparameters
resulting from a parameter sweep performed by the PufferLib team.
To ensure a fair comparison between the single-objective baseline
and our multi-objective variants,
we used the default configuration in \textbf{MOBA} and \textbf{Snake}
and ran a hyperparameter sweep for the MOPPO agent in \textbf{Tetris}.
We selected the best-performing configuration for the MOPPO agent
and maintained these parameters for all subsequent multi-objective extensions
to isolate the algorithmic impact. In all experiments,
the total number of training steps
was kept consistent with the original baselines.

Key experimental settings differ across environments
to match their respective complexities (\zcref[S]{tab:parameters}).
Discount factors are reported as recommended by \citet{castro:2025}.
The discount factor is used during both training and evaluation
to compute discounted returns (\zcref[S]{fig:train_metrics}).

Detailed algorithm-specific hyperparameters
can be found in the configuration files in our repository \citep{puffermo}.
All reported training steps correspond to agent transitions.
Each environment step involves $N_\text{vec} \times N_\text{env} \times N_\text{agents}$ agent transitions,
where $N_\text{vec}$ is the vectorization factor, $N_\text{env}$ is the number of parallel environments,
and $N_\text{agents}$ is the number of agents per environment.
During evaluation, $N_\text{vec} = 1$.

\begin{figure}[!h]
\centering
    \begin{minipage}[t]{0.32\textwidth}
        \centering
        \includegraphics[width=\linewidth]{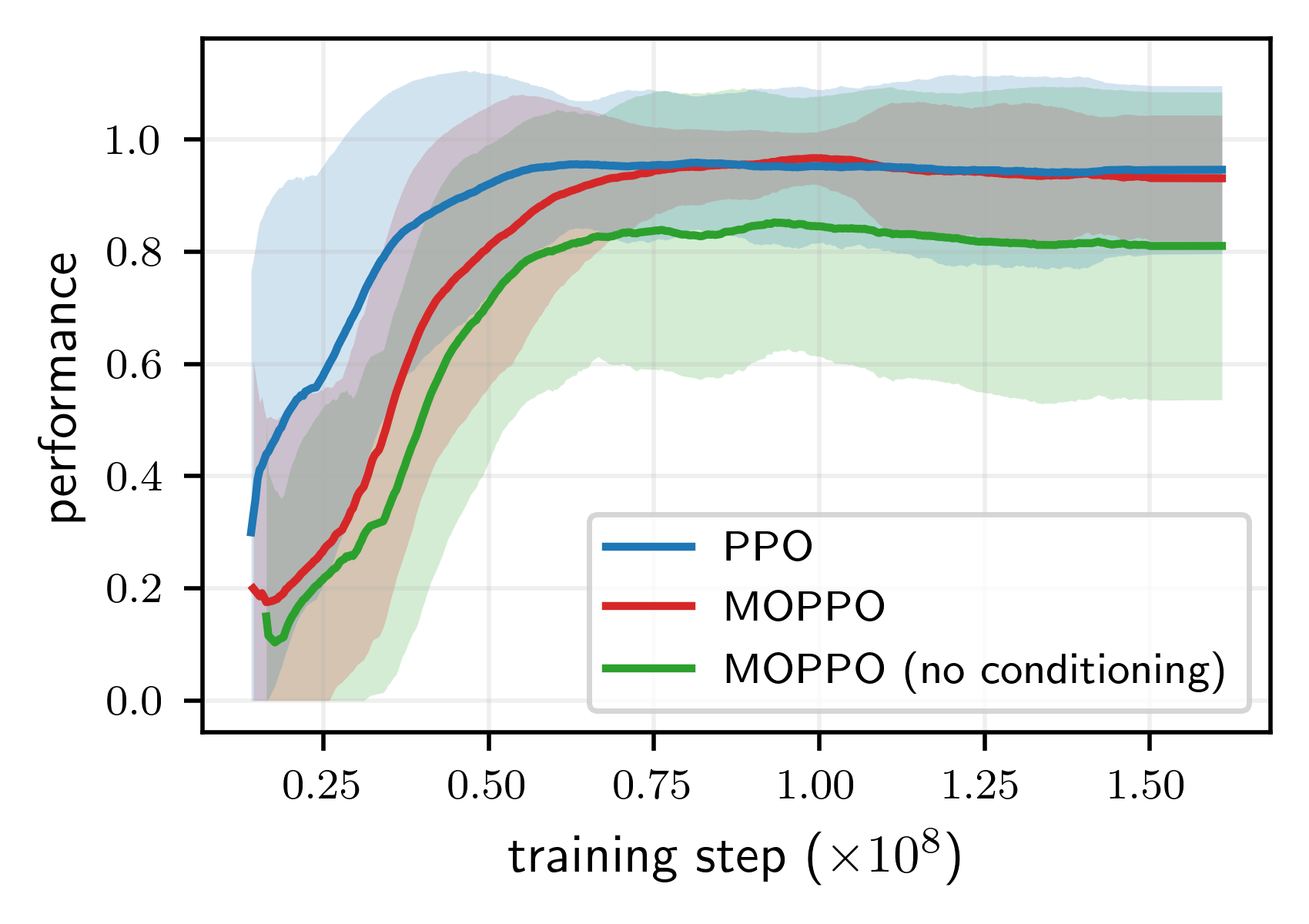}
    \end{minipage}
    \hfill
    \begin{minipage}[t]{0.32\textwidth}
        \centering
        \includegraphics[width=\linewidth]{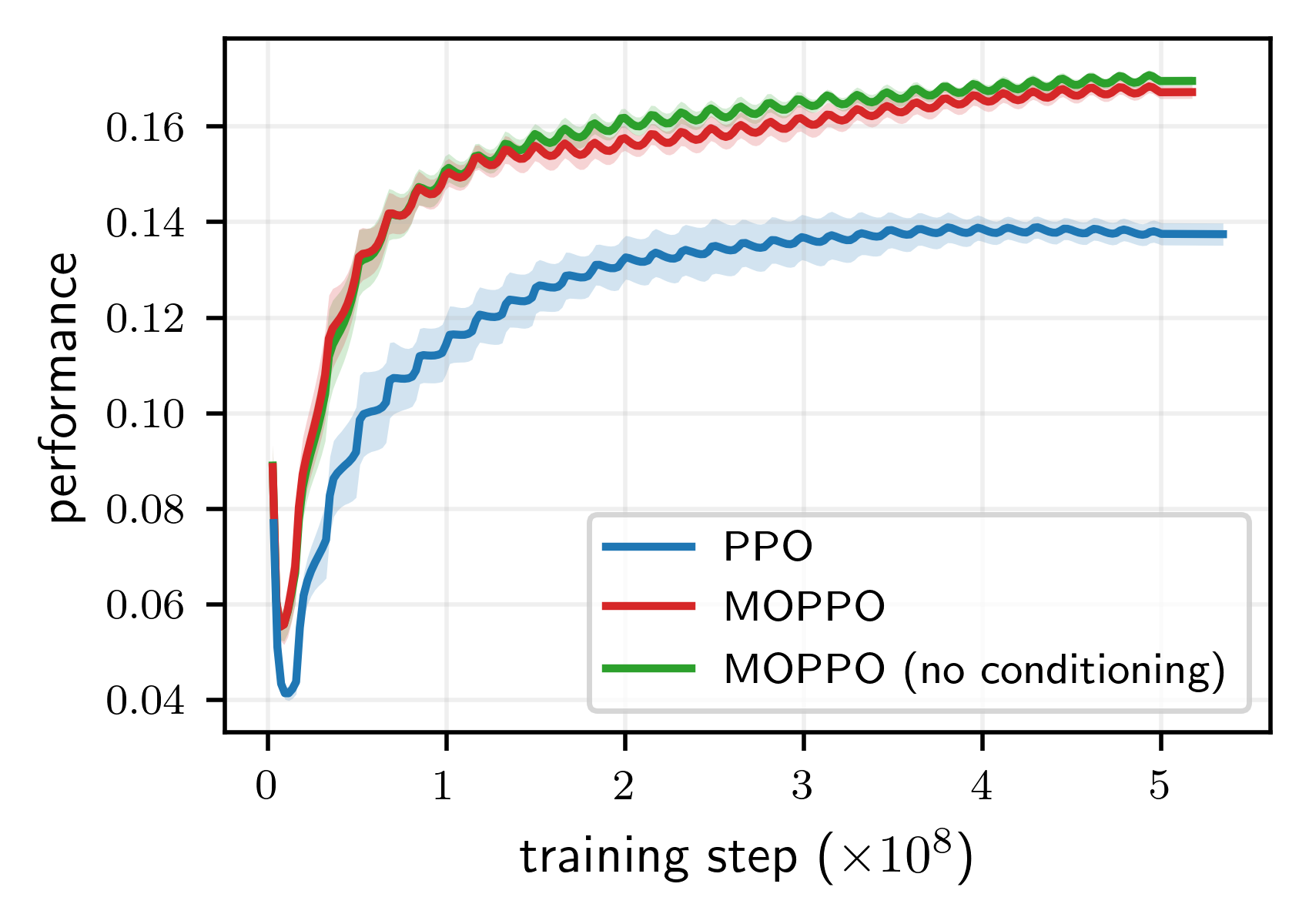}
    \end{minipage}
    \hfill
    \begin{minipage}[t]{0.32\textwidth}
        \centering
        \includegraphics[width=\linewidth]{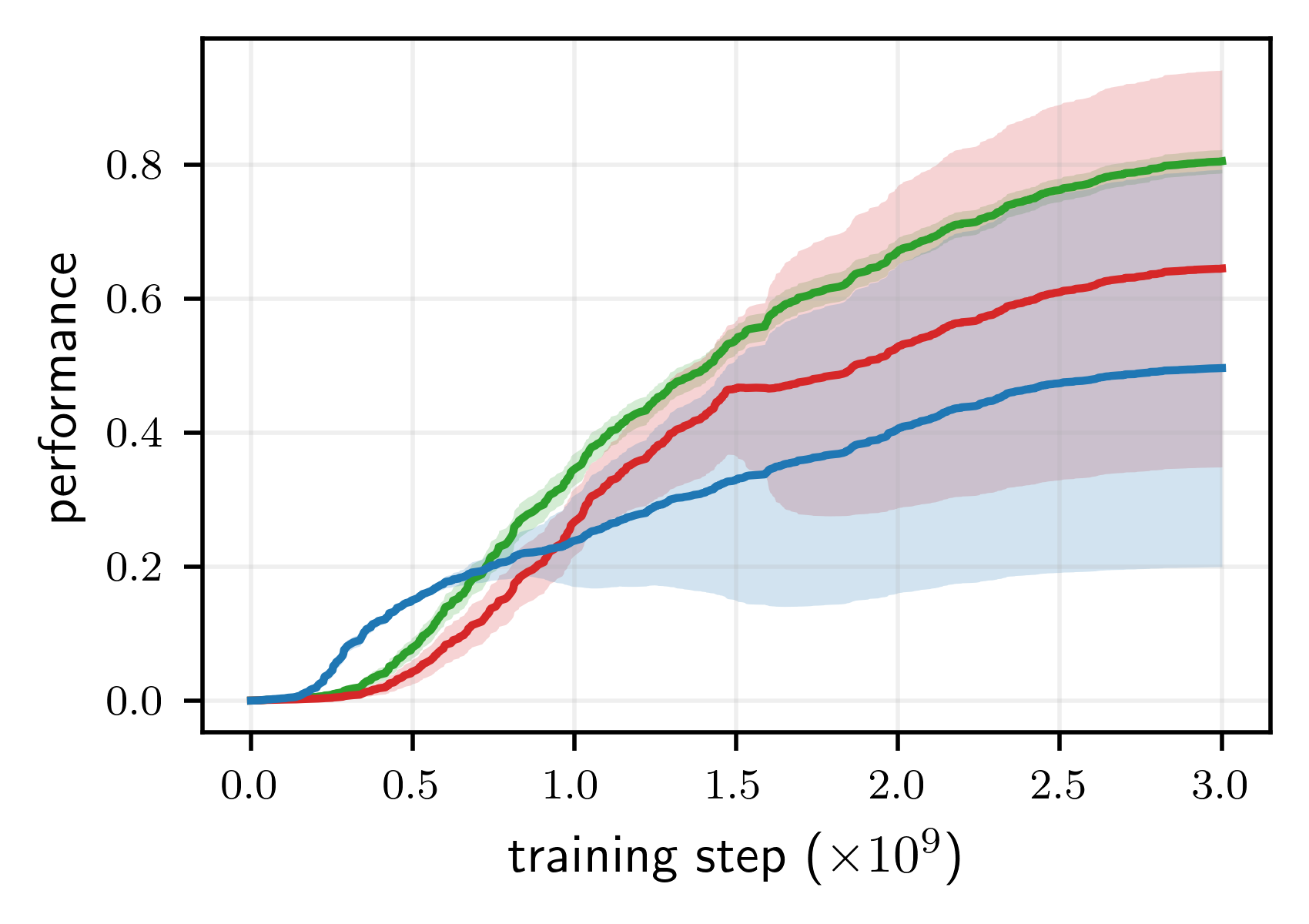}
    \end{minipage}
    \\ \vspace{1em}
    \begin{minipage}[t]{0.32\textwidth}
        \centering
        \includegraphics[width=\linewidth]{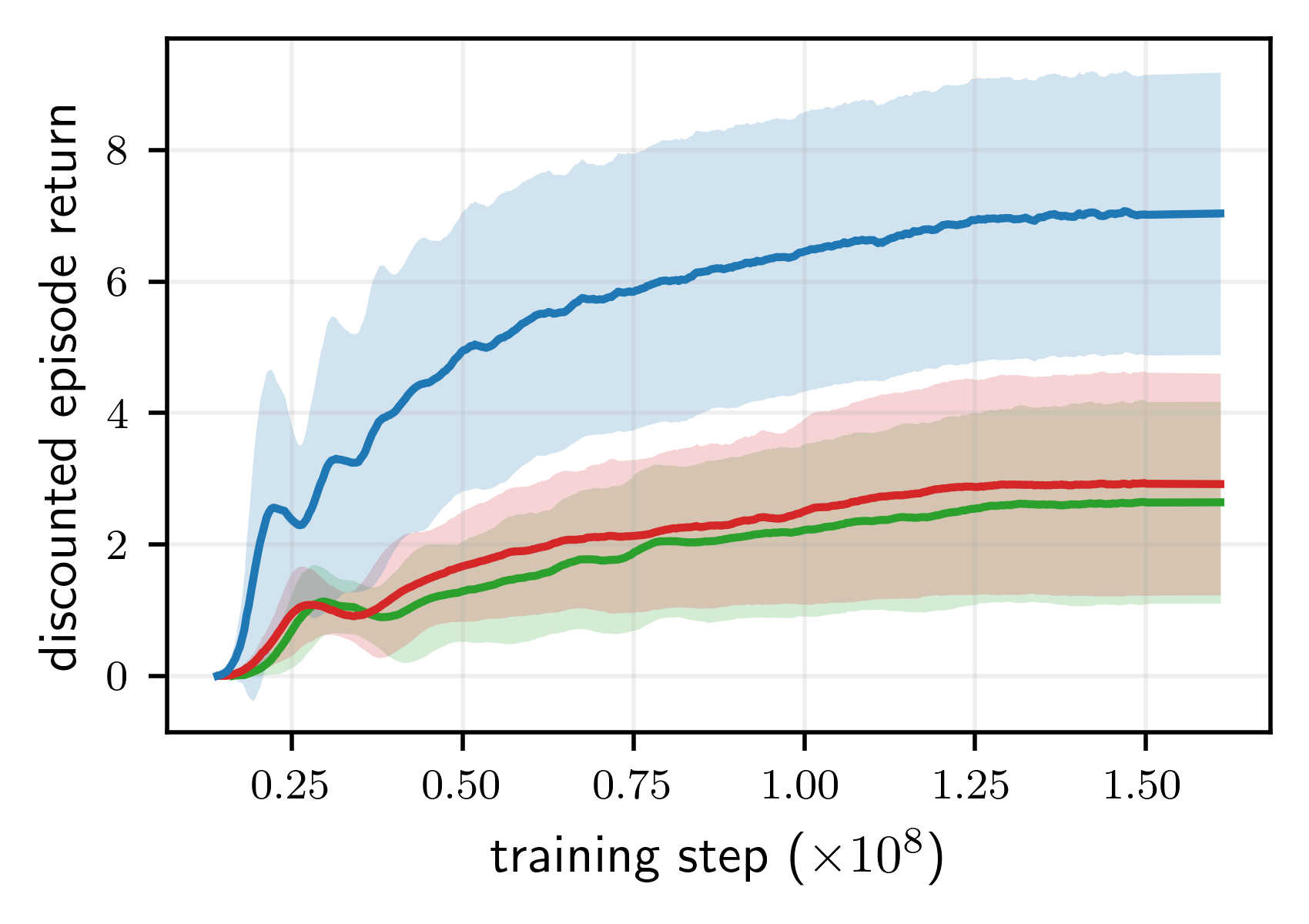}
        \vspace{-1em}
        \par MOBA
    \end{minipage}
    \hfill
    \begin{minipage}[t]{0.32\textwidth}
        \centering
        \includegraphics[width=\linewidth]{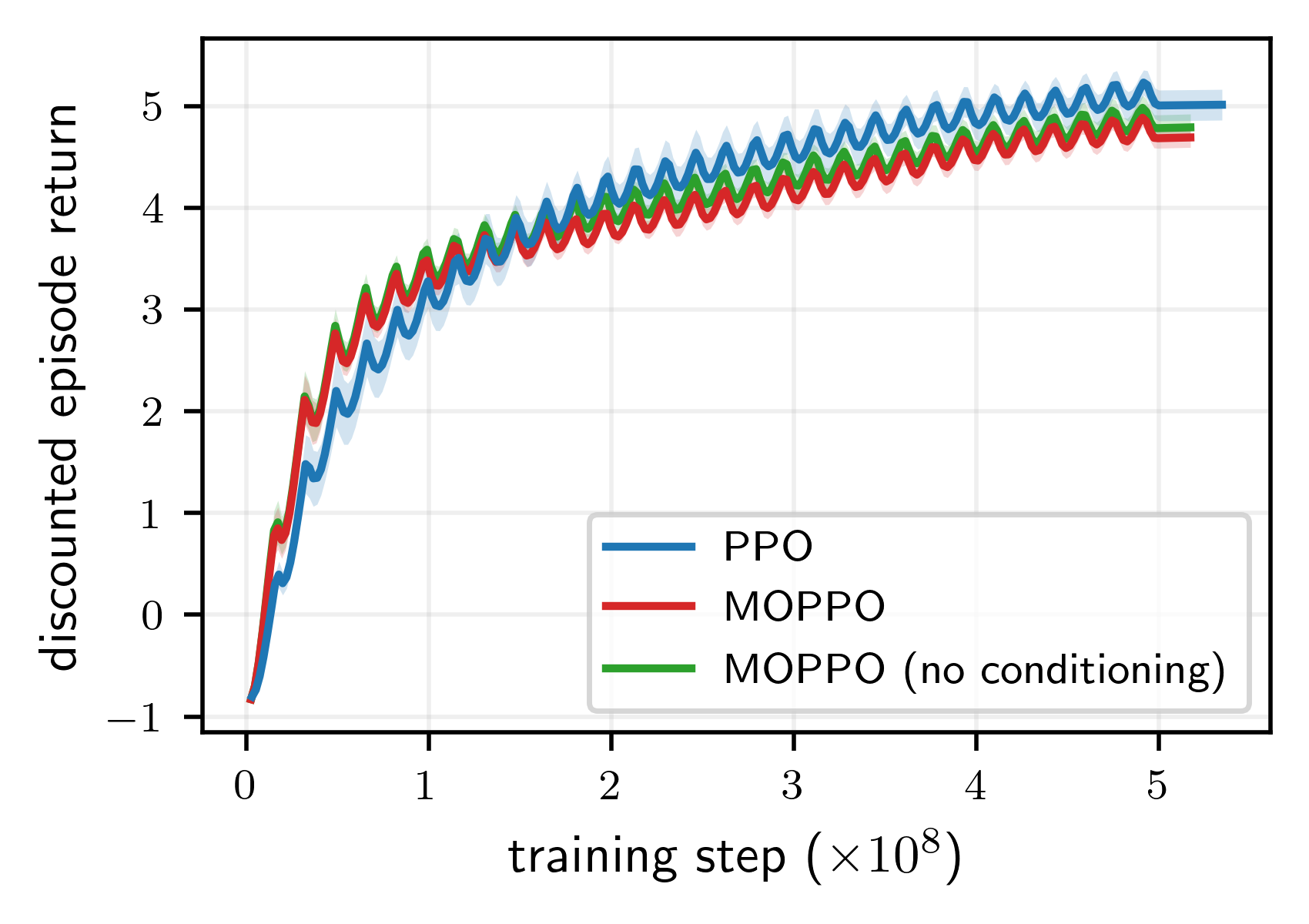}
        \vspace{-1em}
        \par Snake
    \end{minipage}
    \hfill
    \begin{minipage}[t]{0.32\textwidth}
        \centering
        \includegraphics[width=\linewidth]{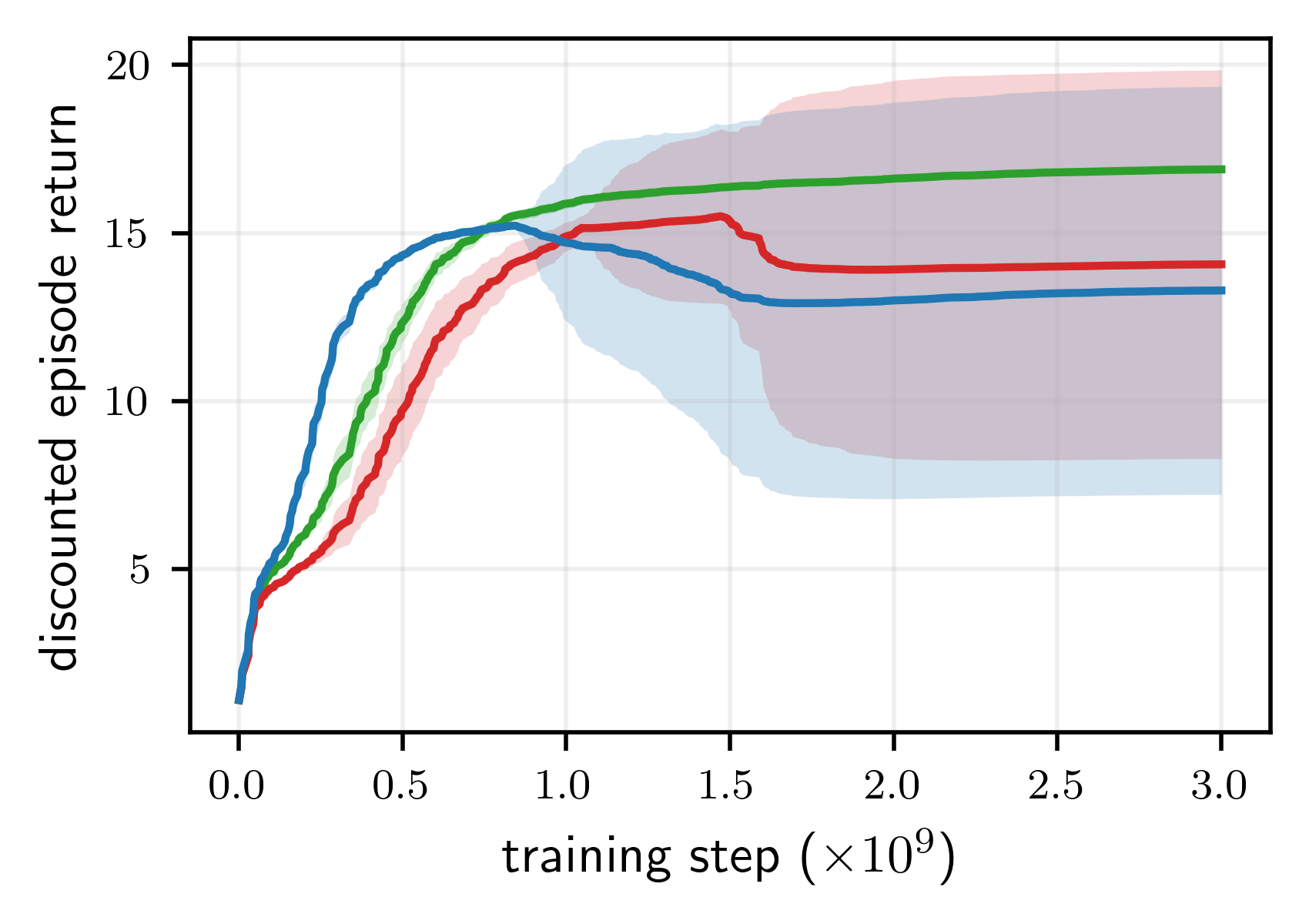}
        \vspace{-1em}
        \par Tetris
    \end{minipage}
\caption{
    MOPPO achieves training performance comparable to PPO,
    validating the multi-objective extension.
    Means are smoothed with an exponential moving average ($\alpha = 0.95$).
    Shaded areas represent one standard deviation.
    }
\label{fig:train_metrics}
\end{figure}

\begin{table}
\renewcommand{\arraystretch}{1.3}
\setlength{\tabcolsep}{6pt}
\centering
\begin{tabular}{l c c c l}
\textbf{Parameter}         & \textbf{MOBA}     & \textbf{Snake}    & \textbf{Tetris}   & \textbf{Description} \\
\multicolumn{5}{@{}l}{\textit{Training}} \\
Total agent steps          & $1.5 \times 10^8$ & $5 \times 10^8$   & $3 \times 10^9$   & Agent transitions for training \\
Train time                 & $\approx 4$ min   & $\approx 3$ min   & $\approx 17$ min  & Wall-clock time per run \\
\multicolumn{5}{@{}l}{\textit{Evaluation}} \\
Eval episodes              & 128               & 4                 & 2048              & Episodes averaged per solution \\
Episode horizon            & $5 \times 10^3$   & $3 \times 10^3$   & $3 \times 10^3$   & Max steps per episode \\
\multicolumn{5}{@{}l}{\textit{Common}} \\
Discount factor ($\gamma$) & 0.995             & 0.9997            & 0.995             & Reward discounting (train, eval) \\
\multicolumn{5}{@{}l}{\textit{Environment}} \\
Agents per environment     & 5                 & 256               & 1                 & Concurrent agents \\
\end{tabular}
\caption{
    Training and evaluation settings vary across environments
    to match their respective complexities.
    Detailed hyperparameters are available in the repository \citep{puffermo}.
    }
\label{tab:parameters}
\end{table}

\subsection{Evaluation metric computation}

\label{sec:metric_computation}

All metrics are computed using a single Python script available in the PufferMO repository \citep{puffermo},
following standard practices, which are briefly described here for completeness.
HV, SP, and CS are susceptible to artifacts from differing return scales across objectives,
so they are computed on min-max normalized returns, with the normalization range derived jointly across all algorithms within each environment.
HV is computed using \texttt{pymoo} on the non-dominated front, with the reference point placed $0.1$ beyond the nadir \citep{wang:2023}.
EU and CS use the full set of returns rather than the Pareto front approximation,
with the 30 evaluation weight vectors serving as a finite surrogate for the uniform distribution over $\Delta^\Nobjectives$.

Spearman's per-objective correlation coefficient $\rho^d$
is computed using the built-in function
packaged with \texttt{scipy} (\texttt{scipy.stats.spearmanr})
on raw (non-normalized) returns paired
with the corresponding conditioning weights.
When either variable has near-zero standard deviation,
the coefficient is undefined and omitted.
Statistical significance is assessed
via the two-sided $p$-value;
results with $p < .001$ are marked
with an asterisk in \zcref[S]{tab:evaluation}.

\end{document}